%% file: main.tex
\documentclass{article}
\usepackage{preprint_style,times}

\usepackage{amsmath,amssymb}
\usepackage{graphicx}
\usepackage[caption=false]{subfig}
\usepackage{booktabs}
\usepackage{tabularx}
\usepackage{array}
\usepackage{multirow}
\usepackage{longtable}
\usepackage{wrapfig}
\usepackage{enumitem}
\usepackage{algorithm}
\usepackage{algpseudocode}
\usepackage{tikz}
\usepackage{colortbl}
\tcbuselibrary{skins,listings}
\usepackage{hyperref}
\usepackage{url}
\usepackage[nameinlink,capitalize,noabbrev]{cleveref}

\definecolor{eliteBoxBg}{HTML}{F2F2F2}
\definecolor{eliteBoxBorder}{HTML}{A6A6A6}
\definecolor{eliteBoxText}{HTML}{4A4A4A}

\newcommand{\elitebox}[1]{%
  \tikz[baseline=(elite.base)]{%
    \node[
      draw=eliteBoxBorder,
      fill=eliteBoxBg,
      text=eliteBoxText,
      line width=0.5pt,
      rounded corners=2pt,
      minimum width=13pt,
      minimum height=13pt,
      inner sep=0pt,
      font=\sffamily\scriptsize\bfseries
    ] (elite) {#1};%
  }%
}

\newcommand{\eliteM}{\elitebox{M}}
\newcommand{\eliteP}{\elitebox{P}}
\newcommand{\eliteT}{\elitebox{T}}
\newcommand{\eliteF}{\elitebox{F}}
\newcommand{\eliteA}{\elitebox{A}}

\newcommand{\elitegap}{%
  \hskip 2pt plus 1pt minus 0.5pt\allowbreak
}

\definecolor{accent}{HTML}{2A78D6}
\definecolor{ink}{HTML}{0B0B0B}
\definecolor{panelgray}{HTML}{F4F4F2}

\newcommand{\targettag}[1]{%
  \tcbox[
    on line,
    arc=2.5pt,
    boxsep=0pt,
    left=3.2pt,
    right=3.2pt,
    top=1.5pt,
    bottom=1.5pt,
    boxrule=0.5pt,
    colback=accent,
    colframe=accent,
    coltext=white,
    fontupper=\sffamily\scriptsize\bfseries
  ]{#1}%
}

\definecolor{schemaBg}{HTML}{EAF2FB}
\definecolor{schemaFrame}{HTML}{8BB4E3}
\definecolor{schemaText}{HTML}{275F99}

\definecolor{extractBg}{HTML}{EAF7EE}
\definecolor{extractFrame}{HTML}{8BC79C}
\definecolor{extractText}{HTML}{2D6B40}

\definecolor{planBg}{HTML}{FFF4E3}
\definecolor{planFrame}{HTML}{DDB36B}
\definecolor{planText}{HTML}{815A1C}

\definecolor{answerBg}{HTML}{F3ECFA}
\definecolor{answerFrame}{HTML}{B499D1}
\definecolor{answerText}{HTML}{654783}

\definecolor{indexBg}{HTML}{FCEDEF}
\definecolor{indexFrame}{HTML}{D99BA3}
\definecolor{indexText}{HTML}{89444D}

\definecolor{retrieveBg}{HTML}{E9F6F6}
\definecolor{retrieveFrame}{HTML}{86BDBD}
\definecolor{retrieveText}{HTML}{2F7070}

\definecolor{helperBg}{HTML}{F1F2F3}
\definecolor{helperFrame}{HTML}{A9AFB5}
\definecolor{helperText}{HTML}{555B61}

\newcommand{\compcolor}[4]{%
  \tcbox[
    on line,
    arc=2pt,
    boxsep=0pt,
    left=2.3pt,
    right=2.3pt,
    top=1.2pt,
    bottom=1.2pt,
    boxrule=0.45pt,
    colback=#2,
    colframe=#3,
    coltext=#4,
    fontupper=\ttfamily\tiny
  ]{#1}%
}

\newcommand{\schema}{\compcolor{schema}{schemaBg}{schemaFrame}{schemaText}}
\newcommand{\extract}{\compcolor{extraction}{extractBg}{extractFrame}{extractText}}
\newcommand{\plan}{\compcolor{planning}{planBg}{planFrame}{planText}}
\newcommand{\answer}{\compcolor{answering}{answerBg}{answerFrame}{answerText}}
\newcommand{\indexc}{\compcolor{indexing}{indexBg}{indexFrame}{indexText}}
\newcommand{\retrievec}{\compcolor{retrieval}{retrieveBg}{retrieveFrame}{retrieveText}}
\newcommand{\helpers}{\compcolor{helpers}{helperBg}{helperFrame}{helperText}}

\newcommand{\topcell}[1]{%
  \begin{minipage}[t]{\linewidth}
  \vspace{0pt}%
  \raggedright
  #1%
  \end{minipage}%
}

\newcommand{\componentsep}{%
  \hspace{1.5pt plus 0.5pt}%
  \allowbreak
}

\newcommand{\componentlabels}[1]{%
  \par\vspace{3pt}%
  \noindent
  \begingroup
  \raggedright
  \setlength{\emergencystretch}{1em}%
  #1\par
  \endgroup
}

\hypersetup{
  pdftitle={Capability-Driven Self-Evolution of Agent Memory},
  pdfauthor={Yaoqi Chen, Yuru Feng, Qianxi Zhang, Baotong Lu, Jianan Lu, Zhirui Wang, Shusen Xu, Zewen Jin, Zengzhong Li, Cheng Li, Qi Chen}
}

\newcommand{\method}{\mbox{\textsc{PrisMem}}}
\newcommand{\mstar}{M$^{\star}$}
\newcommand{\caps}{\mathcal{C}}
\newcommand{\pool}{\mathcal{P}}
\newcolumntype{Y}{>{\raggedright\arraybackslash}X}
\DeclareMathOperator*{\argmax}{arg\,max}

\title{Capability-Driven Self-Evolution of Agent Memory}
\author{
\textbf{Yaoqi Chen$^{1,2}$\thanks{Equal contribution.}, Yuru Feng$^{2,3}$\footnotemark[1],
Qianxi Zhang$^{2}$, Baotong Lu$^{2}$,} \\
\textbf{Jianan Lu$^{2}$, Zhirui Wang$^{2}$, Shusen Xu$^{2}$, Zewen Jin$^{1}$,} \\
\textbf{Zengzhong Li$^{2}$, Cheng Li$^{1}$, Qi Chen$^{2}$} \\[6pt]
$^{1}$University of Science and Technology of China \\
$^{2}$Microsoft \quad $^{3}$University of California, San Diego
}

\begin{document}
\maketitle

\begin{abstract}
Memory self-evolution uses task feedback to iteratively improve executable memory programs that store and retrieve information from past interactions.
Existing approaches typically adopt holistic evolution, deriving revision directions from mixed feedback and judging progress by overall performance.
This can obscure optimization directions and hide capability-specific gains offset by regressions elsewhere, leaving promising directions underexplored.
We introduce \emph{capability-driven evolution}, which extends search guidance from overall performance to individual capability dimensions, preserving promising revisions and expanding exploration beyond the boundaries of holistic evolution.
We propose \method{}, which uses dependency-aware capability selection to prioritize targets with potential cross-capability benefits and history-guided diagnosis to refine capability specialists.
Trace-guided integration compares evaluated programs on paired differential cases, using their behavioral differences to consolidate complementary gains into a unified memory program.
Experiments show that \method{} outperforms the strongest baselines by 10.54 and 7.83 percentage points on BEAM-1M and LongMemEval-M, respectively, demonstrating its effectiveness on million-token histories. 
\end{abstract}

\input{sections/introduction}
\input{sections/related_work}
\input{sections/method}
\input{sections/experiments}
\input{sections/conclusion}

\bibliography{references}
\bibliographystyle{ref_style}

\clearpage
\appendix
\input{sections/appendix}
\end{document}

%% file: sections/introduction.tex
\section{Introduction}
\label{sec:intro}
Memory has become a key component of large language model (LLM) agents, enabling them to retain and leverage information from past interactions~\citep{packer2023memgpt, zhong2024memorybank,kang2025memoryos,salama2025meminsight}. These interactions contain diverse information, including explicit facts, implicit preferences and dynamically changing states. Moreover, different requests call for diverse capabilities, such as factual retrieval, temporal tracking, preference recognition, and cross-session synthesis~\citep{maharana2024locomo,wu2025longmemeval,tavakoli2026beam}. Effective memory is therefore inherently multidimensional, requiring a memory program to support distinct yet interconnected capabilities across diverse tasks and evolving interactions~\citep{xu2025amem,rasmussen2025zep, personaagent, salama2025meminsight,ye2026memweaver}. 

\begin{figure}[t!]
  \centering
  \includegraphics[width=.99\linewidth]{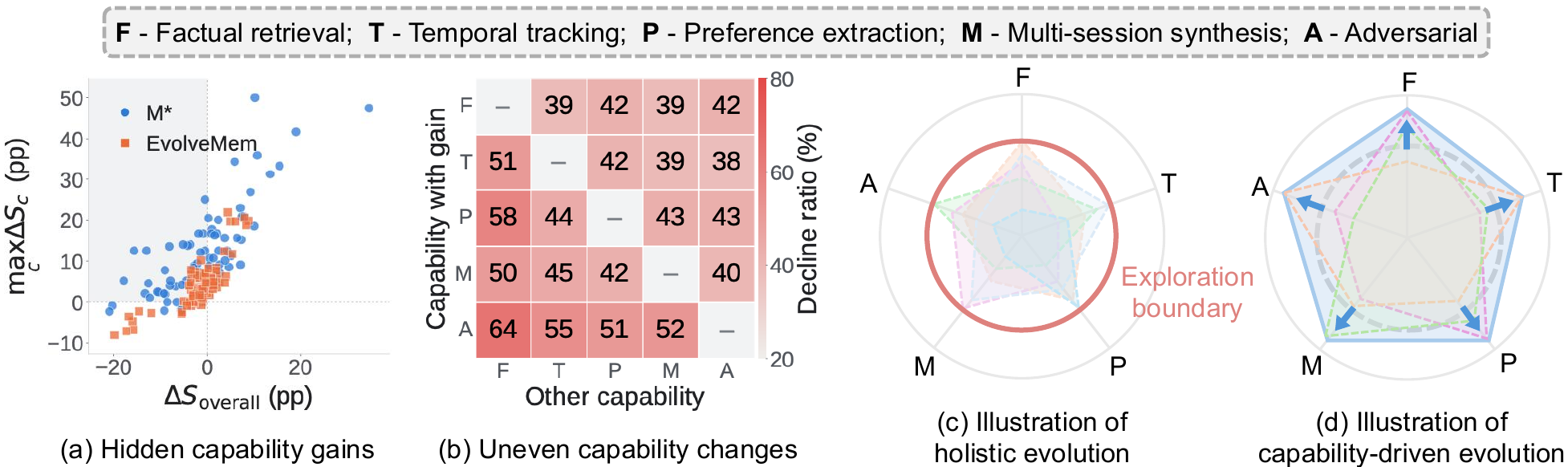}
  \caption{\textbf{Motivation for capability-driven evolution.} Analysis of \mstar{} and EvolveMem evolution traces on BEAM. (a) Overall and maximum capability-level score changes for each revision. Many revisions improve at least one capability despite lower overall scores (upper-left region). (b) Among revisions that improve one capability (row), the percentage for which another capability's score decreases (column). (c) Holistic evolution confines exploration to a limited region. (d) Capability-driven evolution enables broader exploration through explicit capability directions.}
  \label{fig:intro}
\end{figure}

Self-evolving memory offers a promising solution by representing agent memory as an executable program, evaluating it on training tasks, and iteratively refining its implementation by an LLM based on task feedback~\citep{pan2026mstar,liu2026evolvemem,zhang2025memevolve}. A key challenge is identifying useful revision directions and determining which improvements are worth pursuing. Existing methods generally adopt \emph{holistic evolution}, deriving optimization directions from mixed feedback across memory capabilities and judging progress mainly by overall task performance. Such compressed scalar guidance can confine the search to a limited region of capability space, with overall performance plateauing while some capability directions remain underexplored (Fig.~\ref{fig:intro}(c)).

This limitation arises from two issues: \textbf{ambiguous optimization directions} and \textbf{hidden capability gains}. First, mixed feedback combines failures from tasks requiring different capabilities, which may suggest multiple possible revisions for the same memory component with different effects. For example, broader retrieval may improve multi-session reasoning but increase over-personalization by introducing irrelevant personal memories~\citep{hu2026opbench,pulipaka2026persistbench, memcalib}. Our analysis in Fig.~\ref{fig:intro}(b) shows that such uneven effects are common. Before these directions are implemented and evaluated, their actual effects across capabilities remain unknown, making it difficult to determine which direction to pursue for overall improvement.

Second, a revision's gains on one capability can be offset by regressions in others, leaving the overall score unchanged or even lower. Fig.~\ref{fig:intro}(a) shows that 80.5\% of revisions without an overall gain still improve at least one capability. 
This causes useful mechanisms to be overlooked before their benefits can be further refined, integrated, and reflected in overall performance.
This issue is particularly pronounced near a performance plateau, where marginal gains are easily masked, further confining the search around the current solution.

Our key insight is to elevate the evolution space from a single overall-performance dimension to multiple capability-level dimensions. This motivates \emph{capability-driven evolution}:  Capability-specific feedback suggests focused diagnosis and guides explicit directions, while capability-level evaluation preserves promising revisions even without overall gains, enabling further refinement and broader exploration along individual capability dimensions (Fig.~\ref{fig:intro}(d)). Crucially, this process also turns broader optimization directions into evaluated memory programs whose implementations and execution results reveal how differently these programs behave on the same tasks. These references provide guidance for combining useful mechanisms into a stronger unified memory program, rather than choosing among unexplored directions from mixed feedback.

We propose \method{}, which decomposes holistic memory evolution into capability-specific directions, analogous to a prism separating light. 
To achieve broader gains from each capability refinement step, \emph{dependency-aware capability selection} exploits dependencies among capabilities to prioritize those whose improvement can also benefit others. Refinement starts from the best-performing program for the selected capability (specialist), allowing local gains to develop even when obscured by the overall score and expanding the search boundary.
\emph{History-guided diagnosis} then identifies informative failures that past programs handled well, while discounting repeatedly examined cases and capability combinations to broaden diagnostic coverage and uncover further revision opportunities.
Finally, \emph{trace-guided integration} consolidates capability specialists evolved along different directions. Using \emph{paired differential cases}, it compares the base program with each specialist to expose their behavioral differences and guide how complementary gains should be incorporated. This translates broader capability-level exploration into a stronger unified memory program.

Our contributions are: (1) We introduce capability-driven evolution, a new paradigm for memory self-evolution that lifts search guidance to the capability dimension, expanding the exploration boundary. (2) We propose \method{}, which combines dependency-aware capability selection, history-guided diagnosis, and trace-guided integration to develop capability-level advances into a stronger memory program. (3) We evolve on shorter-context splits of BEAM and LongMemEval and test on their 1M-token versions, achieving improvements of 7.83--10.54 percentage points over the best-performing baselines and demonstrating effective generalization to million-token histories.

%% file: sections/related_work.tex
\section{Background and Related Work}
\label{sec:related}

\paragraph{Diverse Memory Capabilities.}
Agent memory enables agents to retain and reuse interaction history for reasoning, decision-making, and long-term interaction~\citep{park2023generative,packer2023memgpt}. 
Diverse historical information and user needs impose heterogeneous requirements on memory systems, where design choices that benefit one capability may adversely affect others.
For instance, personalized assistance requires leveraging a user's preferences, while general information requests should remain unaffected by unrelated personal details~\citep{zhong2024memorybank,hu2026opbench}.
Appendix~\ref{sec:capability_revision_cases} provides more examples.
These diverse requirements motivate different approaches to memory design, including \emph{static} and \emph{self-evolving} approaches.

\paragraph{Static Agent Memory.}
Static memory systems use predefined structures and fixed strategies for storing and retrieving information, with manually designed mechanisms for different memory capabilities.
Mem0~\citep{chhikara2025mem0} and SimpleMem~\citep{liu2026simplemem} emphasize preserving and consolidating salient information for \emph{factual retrieval}.
Temporal context receives particular attention in Zep~\citep{rasmussen2025zep}, Hindsight~\citep{latimer2025hindsight} and TiMem~\citep{li2026timem}, supporting \emph{temporal tracking} of events and changing states.
MemoryBank~\citep{zhong2024memorybank}, MemInsight~\citep{salama2025meminsight}, MemoryOS~\citep{kang2025memoryos}, O-Mem~\citep{wang2025omem}, and EverMemOS~\citep{hu2026evermemos} draw on accumulated user information to support \emph{preference extraction} and personalized responses.
A-MEM~\citep{xu2025amem}, GAM~\citep{yan2025gam}, MRAgent~\citep{ji2026mragent}, MemWeaver~\citep{ye2026memweaver}, and Amory~\citep{zhou2026amory} emphasize connecting and combining information across interactions for \emph{multi-session synthesis}.

\paragraph{Self-Evolving Agent Memory.}
Self-evolving memory methods use task feedback to iteratively refine memory structures and management strategies by an LLM, finally producing a fixed memory program. 
EvolveMem~\citep{liu2026evolvemem} focuses on configuration-level optimization of retrieval strategies, while \mstar{}~\citep{pan2026mstar}, MemEvolve~\citep{zhang2025memevolve}
and MemPro~\citep{liu2026mempro} 
pursue program-level evolution of full memory structures and processing logic.
These methods rely on \emph{holistic evolution} driven by overall scores. This mixed feedback can blur optimization directions, while aggregate scores can mask capability gains that are offset by regressions elsewhere, leading to stagnant evolution.

%% file: sections/method.tex
\section{\method{}: Capability-Driven Memory Evolution}
\label{sec:method}

\paragraph{Problem Formulation.}
\method{} aims to improve how an agent stores and retrieves conversational history by optimizing its executable memory program $P$.
Let $x=(\mathcal{H},q,y,E)$ denote a task instance with history $\mathcal{H}$, question $q$, reference answer $y$, and reference evidence $E$.
The program constructs memory from $\mathcal{H}$ and retrieves content for answering $q$, receiving a task score $s(P;x)\in[0,1]$.
We denote the average score of $P$ over a task set $\mathcal{S}$ as $s(P;\mathcal{S})$. The task set is split into $\mathcal{S}_{\mathrm{train}}$, $\mathcal{S}_{\mathrm{valid}}$, and $\mathcal{S}_{\mathrm{test}}$.
Memory self-evolution iteratively evaluates $P$ on $\mathcal{S}_{\mathrm{train}}$, leverages an LLM to diagnose problems and refine $P$, and pursues improvements on $\mathcal{S}_{\mathrm{valid}}$. After reaching maximum iterations, the best-performing program is frozen for evaluation on $\mathcal{S}_{\mathrm{test}}$.

\paragraph{Fine-grained Memory Interface.}
\label{sec:interface}
The memory interface determines the scope within which the memory program can be modified during self-evolution. We decompose the memory program into five components: \emph{Extraction}, \emph{Indexing}, \emph{Planning}, \emph{Retrieval}, and \emph{Answer}. Compared with the commonly used ``add'' and ``retrieve'' abstraction in agent memory systems, this decomposition provides more localized modification boundaries, helping the coding model identify which stage is responsible for a failure and revise the corresponding component more precisely.

\paragraph{Capability Set for Our Study.}
\label{sec:formulation}
\method{} uses capabilities to organize evolution and guide program refinement. In our experiments, we instantiate the framework with five representative agent memory capabilities, denoted by $\caps$ (\cref{tab:vocabulary}). We select these capabilities through a broad review of widely used memory benchmarks~\citep{maharana2024locomo,wu2025longmemeval,tavakoli2026beam} and existing memory systems~\citep{zhong2024memorybank,rasmussen2025zep,chhikara2025mem0,xu2025amem}, focusing on recurring memory requirements across these settings~\footnote{\Cref{tab:benchmark_capabilities} in Appendix illustrates how different question types from some benchmarks relate to the chosen capabilities. Hereafter, we use \textbf{F}, \textbf{T}, \textbf{P}, \textbf{M}, and \textbf{A} as shorthand for the full capability names.}. Each capability captures a distinct and essential memory requirement.

\begin{table}[t!]
\caption{Five memory capabilities used in our experiments.}
\label{tab:vocabulary}
\centering
\begin{tabularx}{\linewidth}{p{4.0cm}Y}
\toprule
Capability & Memory requirement \\
\midrule
Factual retrieval (\textbf{F}) & Preserve and retrieve explicit facts, values, and other localized details. \\
Temporal tracking (\textbf{T}) & Order events, determine recency, track states and knowledge updates. \\
Preference extraction (\textbf{P}) & Recover and apply user preferences, instructions and constraints. \\
Multi-session synthesis (\textbf{M}) & Aggregate and synthesize evidence across multiple sessions or events. \\
Adversarial (\textbf{A}) & Abstain when memory lacks supporting evidence, or cautiously identify and resolve contradictory or misleading evidence. \\
\bottomrule
\end{tabularx}
\end{table}

We offline annotate tasks in $\mathcal{S}_{\mathrm{train}}$ and $\mathcal{S}_{\mathrm{valid}}$ using an LLM to assign each task one primary capability $\ell(x)\in\mathcal{C}$ and up to two secondary capabilities $\sigma(x)\subseteq\mathcal{C}\setminus\{\ell(x)\}$. These capability labels organize feedback and guide evolution but are never provided to the memory program during evaluation, ensuring a fair comparison with baselines.
Let $\mathcal{S}^c=\{x\in\mathcal{S}:\ell(x)=c\}$ denote the subset of tasks associated with primary capability $c$.

\subsection{Capability-Driven Evolution}
\label{sec:outer}

\method{} organizes evolution into three stages to expand the search boundary along individual capability dimensions. First, the cold start stage addresses broad weaknesses in the seed program. Second, the capability refinement stage performs \emph{dependency-aware capability selection} to prioritize capabilities and \emph{history-guided diagnosis} to provide focused revisions, while capability-level evaluation preserves promising variations. Finally, \emph{trace-guided integration} consolidates these specialized programs into a stronger, unified memory program. \Cref{fig:realization} illustrates the evolution process, and \Cref{alg:evolution} in Appendix presents the pseudocode.

\subsubsection{Stage I: Cold Start}

The initial seed memory program $P_0$ is deliberately simple, providing a broad search space but potentially leaving basic processing weaknesses across multiple capabilities. Cold start jointly considers capability-wise feedback across all capabilities over $T_1=5$ rounds to quickly establish a balanced base program for subsequent capability refinement. The program with the highest validation score becomes the starting point for the subsequent stages, while the evolution history is retained to guide later capability selection and refinement.

To guide these early revisions, we construct a metric table that aggregates evidence-based diagnostics by capability, preserving capability-specific differences and reducing diagnostic ambiguity. The diagnostics measure semantic similarity between reference evidence and (i) extracted memory units, (ii) content retrieved using the evidence as a query cue, and (iii) content retrieved for the actual question. These metrics assess information preservation during extraction, accessibility through indexing, and retrieval effectiveness. Comparing these capability-by-component patterns helps the model localize broad processing weaknesses and prioritize concrete revisions in early evolution.

\subsubsection{Stage II: Capability Refinement}

Next, we refine individual capabilities of the memory program obtained from cold start stage, retaining valuable capability-specific breakthroughs when they are obscured by overall performance. Let $\pool_t$ denote the set of candidate programs retained up to iteration $t$. For each capability $c$, we track its best-performing program as the specialist $P_t^{\star}(c)$, and denote the globally best program on the complete task set by $G_t$. When a target capability $c_t$ is selected, refinement is initialized from its specialist, i.e., $P_t=P_t^{\star}(c_t)$. Each retained candidate is evaluated on tasks across all capabilities, so a single program can simultaneously serve as the specialist for multiple capabilities.

\begin{figure}[t!]
  \centering
  \includegraphics[width=0.98\linewidth]{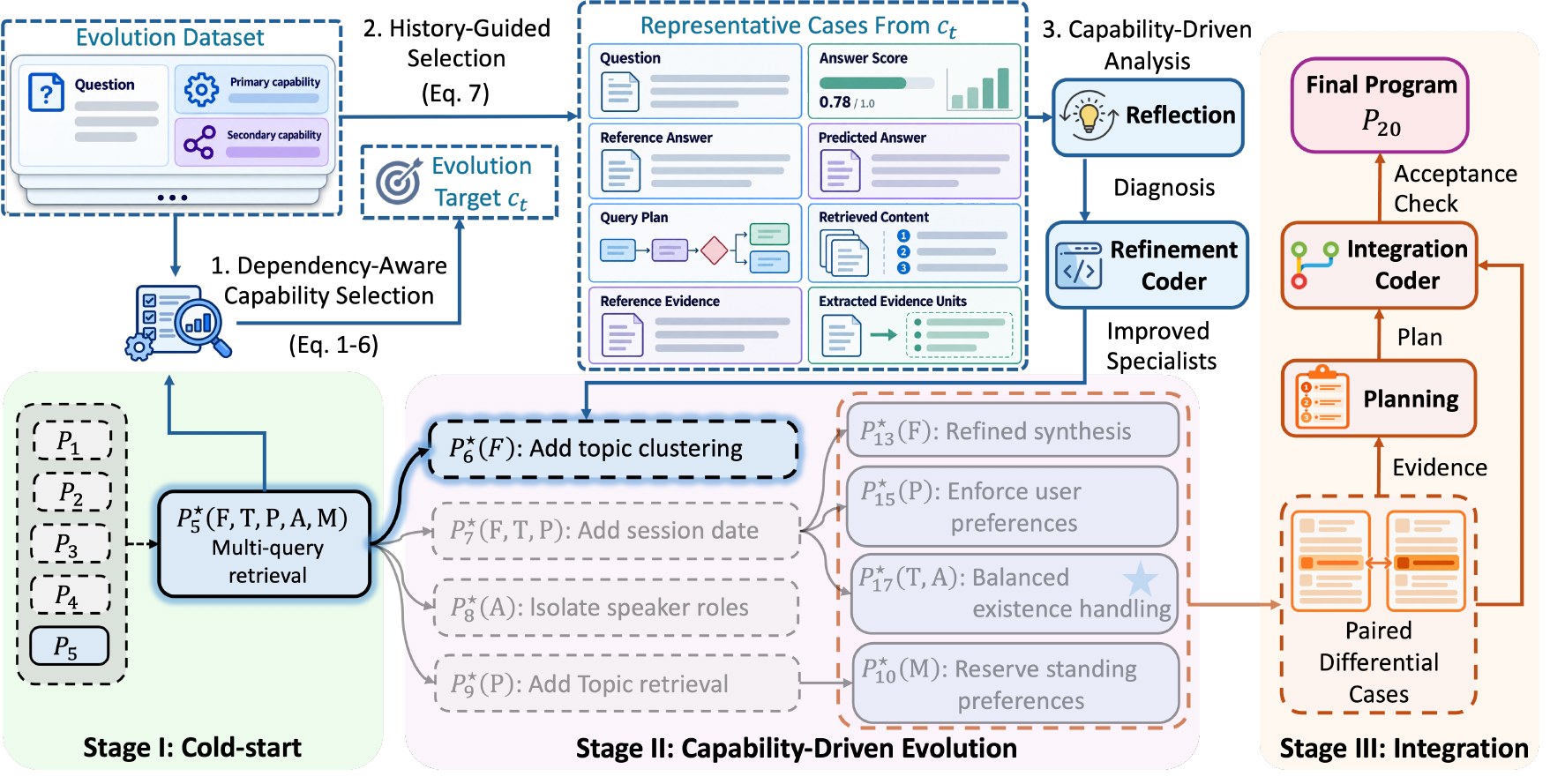}
  \caption{\textbf{Overview of \method{}.} \textit{Stage I: Cold start} quickly establishes a balanced program and initial evolution history. \textit{Stage II: Capability refinement} uses dependency-aware capability selection and history-guided case selection to target capabilities and diagnose informative failures, respectively, refining their specialists through reflection and code revision. \textit{Stage III: Trace-guided integration} compares the base program ($\star$) and other capability specialists on paired differential cases, using their behavioral differences to guide integration into a unified memory program.}
  \label{fig:realization}
\end{figure}

\paragraph{Dependency-Aware Capability Selection.}
Selecting a capability for the next refinement is important because it determines how effectively the limited refinement budget expands the capability boundary. We prioritize capabilities that are weak themselves or appear to be bottlenecks for other capabilities, while downweighting those that have already received substantial refinement budget. A capability may exhibit not only a \textit{direct deficit} on tasks where it is the primary requirement, but also \textit{latent error pressure} on tasks where it appears as a secondary requirement. For example, a deficiency in temporal tracking may affect tasks primarily targeting multi-session synthesis when they also require temporal information.
To quantify this latent pressure, we consider tasks with primary capability $c_p$ and secondary capability $c_s$. Since such co-occurrence does not necessarily imply that deficiencies in $c_s$ affect $c_p$, we first define a structural base-rate prior:
\begin{equation}
  D(c_p\!\rightarrow\!c_s)
  =\frac{\sum_{x\in\mathcal{S}_{\mathrm{train}}^{c_p}}
  \mathbf{1}[c_s\in\sigma(x)]}
  {|\mathcal{S}_{\mathrm{train}}^{c_p}|},
  \label{eq:dependency_prior}
\end{equation}
which measures how often primary-$c_p$ tasks also require $c_s$. At iteration $t$, let
\begin{equation}
  e_{t,c_s}(x)=1-s(P_t^\star(c_s);x),
  \qquad
  \bar e_{t,c_s}(c_p)=\frac{1}{|\mathcal{S}_{\mathrm{train}}^{c_p}|}
  \sum_{x\in\mathcal{S}_{\mathrm{train}}^{c_p}}e_{t,c_s}(x)
  \label{eq:capability_error}
\end{equation}
denote the instance-level and average error of the $c_s$ specialist on primary-$c_p$ tasks. We then measure how disproportionately this error is concentrated on tasks that require $c_s$:
\begin{equation}
  F_{t,c_s}(c_p\!\rightarrow\!c_s)
  =\frac{\sum_{x\in\mathcal{S}_{\mathrm{train}}^{c_p}}
  e_{t,c_s}(x)\mathbf{1}[c_s\in\sigma(x)]}
  {\sum_{x\in\mathcal{S}_{\mathrm{train}}^{c_p}}e_{t,c_s}(x)},
  \label{eq:error_concentration}
\end{equation}
with $F_{t,c_s}(c_p\!\rightarrow\!c_s)=0$ when the denominator vanishes. Intuitively, if $c_s$ is not a bottleneck for $c_p$, its error should be distributed roughly according to the base rate, i.e., $F\approx D$. Thus, an excess gap $F-D>0$ indicates that deficiencies in $c_s$ are disproportionately associated with errors on $c_p$. Scaling this gap by the primary task error yields the incoming error pressure:
\begin{equation}
  I_t(c_p\!\rightarrow\!c_s)
  =\bar e_{t,c_s}(c_p)
  \bigl[F_{t,c_s}(c_p\!\rightarrow\!c_s)-D(c_p\!\rightarrow\!c_s)\bigr]_{+},
  \label{eq:incoming_pressure}
\end{equation}
where $[z]_{+}=\max(z,0)$. For a candidate capability $c$, we combine its direct deficiency with its strongest cross-capability impact:
\begin{equation}
  U_t(c)=\bar e_{t,c}(c)
  +\max_{c_p\in\caps\setminus\{c\}}I_t(c_p\!\rightarrow\!c).
  \label{eq:target_urgency}
\end{equation}
Crucially, using the maximum rather than a sum avoids degree-centrality bias and focuses the scheduler on the most acute structural bottleneck. Finally, to avoid repeatedly allocating the search budget to plateaued capabilities, we discount urgency by the historical allocation count $n_t(c)$:
\begin{equation}
  \rho_t(c)=\frac{U_t(c)}{1+n_t(c)},
  \qquad
  c_t=\argmax_{c\in\caps}\rho_t(c).
  \label{eq:target_priority}
\end{equation}
Thus, a capability $c_t$ is prioritized either for its direct deficiency or for its strongest upstream impact on other capabilities, while its priority naturally decreases with repeated exploration.

\paragraph{History-Guided Diagnosis.}
In later-stage evolution, the memory program is relatively mature, so further improvements require fine-grained diagnosis of individual failures. Inspecting failure cases through their execution results helps pinpoint whether an error arises from noisy extraction, defective memory representation, retrieval failure, or answer misinterpretation, translating behavioral symptoms into actionable code revisions. However, diagnostic cases can be lengthy. We therefore select $K=5$ cases that are severe, historically actionable, and collectively diverse, allowing each refinement step to explore a broader and more valuable set of optimization directions.

Specifically, we first consider training task instances whose primary capability is the selected target $c_t$ and filter out those with scores above $\delta=0.8$, retaining cases that are not fully correct. We then rank the remaining failures by a history-guided priority weight:
\begin{equation}
  w_t(i) = \frac{1 + (1 / |\mathcal{P}_t(c_t)|) \sum_{P \in \mathcal{P}_t(c_t)} s(P; x_i)}{(1 + m_t(i))(1 + m_t(\sigma(x_i)))}.
  \label{eq:case_weight}
\end{equation}
Here, $\mathcal{P}_t(c_t)$ contains valid programs produced in previous iterations that also targeted $c_t$. The numerator measures historical performance; a high historical score for a current failure identifies a regression witness, whose contrast with the current failure helps localize the effects of recent revisions. The denominator discounts cases that have been selected frequently ($m_t(i)$) and secondary-capability combinations that have been repeatedly covered ($m_t(\sigma(x_i))$), promoting diverse coverage and reducing redundant optimization of similar failures. This enables the limited refinement budget to explore broader and more informative optimization directions.

\subsubsection{Stage III: Trace-guided Integration}
\label{sec:merge}
While capability specialists extend the performance boundary along different dimensions, their improvements may remain isolated. The integration stage consolidates these distributed gains into a unified memory program, starting from the overall best program so far and progressively incorporating capability specialists from other directions. A planning model determines the integration order based on the lineage topology, processing specialists from those closest to the base program to those farther away, and produces a step-by-step integration plan. A coding model then executes the plan.

This integration is enabled by the evaluation results of retained programs. Since all programs are evaluated on the same training task set, we can compare revisions on the same cases and trace their effects through the evolution lineage, revealing their capability-specific gains and side effects. We therefore introduce a \textit{paired differential cases} strategy to guide integration. For each capability, we select the training case with the largest score gap between the base program and the specialist being integrated. Comparing their execution results exposes the specialist's behavioral differences, helping the planning model determine how its gains and side effects should be incorporated. 

The resulting plan is executed for up to $T_3=3$ rounds. If integration fails to improve the overall score or causes any capability to decrease by more than $\theta$, the failed implementation and its execution results on the same differential cases are returned to the coding model for another integration. Overall, this process aims to translate complementary capability-specific improvements into stronger overall performance while mitigating regressions across capabilities.

\subsection{Efficient Evolution}
\label{sec:implementations}

Self-evolution repeatedly revises and evaluates memory programs, incurring substantial inference costs. We reduce this cost through \emph{compact diagnostic contexts} and \emph{revision-aware execution reuse}.  First, diagnostic cases provide the information needed to identify program failures, including the question, reference answer and evidence, retrieved content, and model prediction. Full traces can be lengthy, so we construct compact diagnostic contexts by removing redundant content and marking omissions while retaining snippets relevant to the question and answer. This reduces diagnostic-case token cost by 44\% on average while preserving key information for diagnosis.

Second, evaluating each candidate program requires executing the memory pipeline, but revisions may affect only a subset of its components. Using the fine-grained memory interface, we compare each candidate with its parent and reuse outputs from unaffected components. For example, a retrieval-only revision can reuse the constructed memory and skip extraction and indexing. Selectively reusing completed results reduces redundant processing and accelerates program evaluation.

%% file: sections/experiments.tex
\section{Experiments}
\label{sec:experiments}

We evaluate \method{} on four questions:
\textbf{Q1:} How does \method{} perform compared with existing methods?
\textbf{Q2:} What is the evolution cost of \method{}?
\textbf{Q3:} How much does each component contribute to \method{}'s performance?
\textbf{Q4:} Can \method{} expand the capability boundary and integrate the resulting improvements into the final program?

\subsection{Experimental Setup}
\label{sec:experimental_setup}

\paragraph{Benchmarks.}
We evaluate on two widely used agent memory benchmarks, BEAM~\citep{tavakoli2026beam} and LongMemEval~\citep{wu2025longmemeval}, which cover diverse question types over long, multi-session histories. All evolution methods train their memory programs for 20 iterations on the smaller splits, BEAM-100K and LongMemEval-S, and are then tested on the longer splits with different content, BEAM-1M and LongMemEval-M, with over 1 million tokens. Dataset details can be found in Appendix~\ref{sec:dataset_details}. This protocol tests whether the learned design choices generalize to longer contexts.

\paragraph{Baselines.}
We compare \method{} with seven representative agent memory methods, including static and self-evolving approaches. Static baselines include Mem0~\citep{chhikara2025mem0} and A-MEM~\citep{xu2025amem}, which organize and update structured memories; HippoRAG2~\citep{gutierrez2025hipporag2}, which supports memory-enhanced retrieval with graphs; and SimpleMem~\citep{liu2026simplemem} and LightMem~\citep{fang2025lightmem}, which emphasize efficient memory construction and retrieval. Self-evolving methods include \mstar{}~\citep{pan2026mstar}, which evolves executable memory programs through population-based reflective code search, and EvolveMem~\citep{liu2026evolvemem}, which uses failure diagnosis to adapt retrieval infrastructure and answer-generation configurations. Both adopt holistic evolution, using aggregate task performance as the primary signal for exploration.

\paragraph{Models.}
We use Qwen3.8-27B~\citep{qwen3.8-27B} as the task, reflection, coding, and judging model, and Qwen3-Embedding-8B~\citep{qwen3-embedding} as the embedding model.
Inference runs on NVIDIA A100 GPUs~\citep{A100} with vLLM 0.20.0~\citep{kwon2023efficient} under Python 3.12. We also report results with GPT-5.5~\citep{GPT5.5} on BEAM in Appendix~\ref{sec:gpt_results}.

\paragraph{Metrics.}
Task performance is measured by the LLM judge scores provided by each benchmark. Overall score is averaged over all questions, while capability-level scores are averaged over questions grouped by the native question-type mapping in \cref{tab:benchmark_capabilities}. Inference cost is measured by the average number of task LLM input and output tokens per question. For self-evolving methods, we additionally report the token consumption of the whole evolution procedure. Each experiment is run three times, and results are reported as mean $\pm$ standard deviation.

\subsection{Overall Performance (Q1)}
\label{sec:main_results}
 
\Cref{tab:main} shows the main results on two million-token memory benchmarks.
Static methods show different capability strengths: HippoRAG2 leads in factual retrieval, while A-MEM is competitive in preference extraction. Holistic evolution yields only limited gains over static methods, with \mstar{} improving over the strongest static baseline by 0.99 pp on BEAM and EvolveMem by 0.34 pp on LongMemEval. This limited improvement reflects how mixed feedback can obscure optimization directions and aggregate scores can mask capability gains offset by regressions elsewhere.

In contrast, \method{} achieves the best performance in nearly all capability comparisons, improving overall scores over the strongest static and self-evolving baselines by 8.17--11.53 pp and 7.83--10.54 pp, respectively. These results demonstrate the effectiveness of capability-driven evolution: capability refinement expands exploration along individual dimensions and develops useful revisions beyond holistic evolution, while trace-guided integration consolidates these gains into a stronger program. These gains are achieved with token usage comparable to that of other methods.

\method{} also generalizes well across context lengths and underlying models. Programs evolved on BEAM-100K and LongMemEval-S retain their advantages on the longer BEAM-1M and LongMemEval-M splits. With GPT-5.5, \method{} also outperforms the baselines by 6.17--20.41 pp on BEAM-1M (Appendix~\ref{sec:gpt_results}), showing that capability-driven evolution can produce effective memory programs across different underlying models. Additional cross-model and cross-dataset transfer experiments further support the transferability of the evolved programs (Appendix~\ref{sec:transfer_experiments}).

\begin{table}[t]
\caption{Performance on BEAM-1M and LongMemEval-M with Qwen3.8-27B. ``Score'' denotes the benchmark-provided LLM judge score, and ``Avg. \#Tokens'' denotes the task LLM's average token usage per question. All results are reported as mean $\pm$ std. Best results are in \textbf{bold}.}
\label{tab:main}
\centering
\setlength{\tabcolsep}{3pt}
\begin{tabular}{lccccc>{\itshape}cc}
\toprule
\multirow{3}{*}{Method} & \multicolumn{6}{c}{Score (\%) $\uparrow$} & \multirow{3}{*}{\shortstack{Avg. \#Tokens\\(K) $\downarrow$}} \\
\cmidrule(lr){2-7}
& Factual & Temporal & Preference & Multi-session & \multirow{2}{*}{Adversarial} & \multirow{2}{*}{Overall} & \\
& retrieval & tracking & extraction & synthesis & & & \\
\midrule
\multicolumn{8}{l}{\textbf{\textit{BEAM-1M}}} \\
Mem0 & 54.74{\scriptsize $\pm$ 3.06} & 46.51{\scriptsize $\pm$ 0.79} & 60.48{\scriptsize $\pm$ 2.39} & 38.02{\scriptsize $\pm$ 0.63} & 49.11{\scriptsize $\pm$ 1.24} & 48.95{\scriptsize $\pm$ 0.61} & 190.28{\scriptsize $\pm$ 0.20} \\
A-MEM & 56.21{\scriptsize $\pm$ 1.83} & 46.22{\scriptsize $\pm$ 0.62} & 65.39{\scriptsize $\pm$ 0.52} & 53.87{\scriptsize $\pm$ 0.86} & 41.13{\scriptsize $\pm$ 0.51} & 51.56{\scriptsize $\pm$ 0.62} & 428.95{\scriptsize $\pm$ 3.58} \\
HippoRAG2 & 65.24{\scriptsize $\pm$ 1.26} & 47.06{\scriptsize $\pm$ 1.85} & 66.43{\scriptsize $\pm$ 1.57} & 44.83{\scriptsize $\pm$ 0.60} & 44.82{\scriptsize $\pm$ 2.28} & 51.86{\scriptsize $\pm$ 0.87} & 207.70{\scriptsize $\pm$ 0.19} \\
SimpleMem & 48.55{\scriptsize $\pm$ 3.33} & 36.82{\scriptsize $\pm$ 0.47} & 48.85{\scriptsize $\pm$ 1.19} & 21.87{\scriptsize $\pm$ 2.00} & 50.68{\scriptsize $\pm$ 0.40} & 40.18{\scriptsize $\pm$ 0.76} & 95.72{\scriptsize $\pm$ 0.24} \\
LightMem & 46.66{\scriptsize $\pm$ 0.82} & 33.40{\scriptsize $\pm$ 0.65} & 36.64{\scriptsize $\pm$ 0.63} & 16.70{\scriptsize $\pm$ 0.89} & 55.45{\scriptsize $\pm$ 0.94} & 36.44{\scriptsize $\pm$ 0.38} & \textbf{86.20}{\scriptsize $\pm$ \textbf{0.05}} \\
\midrule
\mstar{} & 61.29{\scriptsize $\pm$ 5.14} & 48.98{\scriptsize $\pm$ 5.87} & 71.91{\scriptsize $\pm$ 3.40} & 42.19{\scriptsize $\pm$ 4.88} & 46.02{\scriptsize $\pm$ 7.52} & 52.85{\scriptsize $\pm$ 2.49} & 89.47{\scriptsize $\pm$ 5.11} \\
EvolveMem & 41.80{\scriptsize $\pm$ 4.77} & 39.74{\scriptsize $\pm$ 3.15} & 52.21{\scriptsize $\pm$ 1.53} & 21.67{\scriptsize $\pm$ 3.07} & 45.67{\scriptsize $\pm$ 3.50} & 40.01{\scriptsize $\pm$ 1.46} & 98.40{\scriptsize $\pm$ 4.67} \\
\rowcolor{preprintAbstractBg}
\method{} & \textbf{74.17}{\scriptsize $\pm$ \textbf{4.87}} & \textbf{57.26}{\scriptsize $\pm$ \textbf{2.15}} & \textbf{75.58}{\scriptsize $\pm$ \textbf{3.91}} & \textbf{54.12}{\scriptsize $\pm$ \textbf{3.37}} & \textbf{64.32}{\scriptsize $\pm$ \textbf{6.51}} & \textbf{63.39}{\scriptsize $\pm$ \textbf{1.02}} & 100.59{\scriptsize $\pm$ 5.02} \\
\midrule[\heavyrulewidth]
\multicolumn{8}{l}{\textbf{\textit{LongMemEval-M}}} \\
Mem0 & 69.79{\scriptsize $\pm$ 1.80} & 63.54{\scriptsize $\pm$ 0.90} & 43.33{\scriptsize $\pm$ 0.00} & 63.54{\scriptsize $\pm$ 1.80} & 63.33{\scriptsize $\pm$ 5.77} & 62.50{\scriptsize $\pm$ 0.50} & 5772.67{\scriptsize $\pm$ 4.12} \\
A-MEM & 87.50{\scriptsize $\pm$ 1.56} & 59.90{\scriptsize $\pm$ 0.90} & 44.44{\scriptsize $\pm$ 1.92} & 54.17{\scriptsize $\pm$ 1.80} & \textbf{96.67}{\scriptsize $\pm$ \textbf{5.77}} & 67.33{\scriptsize $\pm$ 0.76} & 8873.92{\scriptsize $\pm$ 14.45} \\
HippoRAG2 & \textbf{89.58}{\scriptsize $\pm$ \textbf{1.80}} & 47.92{\scriptsize $\pm$ 0.90} & 24.44{\scriptsize $\pm$ 1.92} & 40.63{\scriptsize $\pm$ 3.13} & 90.00{\scriptsize $\pm$ 10.00} & 58.67{\scriptsize $\pm$ 1.26} & 6185.48{\scriptsize $\pm$ 5.01} \\
SimpleMem & 82.81{\scriptsize $\pm$ 1.56} & 74.48{\scriptsize $\pm$ 2.39} & 11.11{\scriptsize $\pm$ 1.92} & 67.71{\scriptsize $\pm$ 1.80} & 73.33{\scriptsize $\pm$ 11.55} & 66.50{\scriptsize $\pm$ 0.50} & 2624.68{\scriptsize $\pm$ 7.90} \\
LightMem & 47.92{\scriptsize $\pm$ 0.90} & 74.48{\scriptsize $\pm$ 0.90} & 24.44{\scriptsize $\pm$ 1.92} & 58.33{\scriptsize $\pm$ 1.80} & \textbf{96.67}{\scriptsize $\pm$ \textbf{5.77}} & 57.00{\scriptsize $\pm$ 0.87} & 1888.30{\scriptsize $\pm$ 1.28} \\
\midrule
\mstar{} & 76.04{\scriptsize $\pm$ 5.49} & 47.40{\scriptsize $\pm$ 3.93} & 12.22{\scriptsize $\pm$ 5.09} & 67.71{\scriptsize $\pm$ 4.77} & 93.33{\scriptsize $\pm$ 11.55} & 56.83{\scriptsize $\pm$ 3.21} & \textbf{1528.94}{\scriptsize $\pm$ \textbf{30.34}} \\
EvolveMem & 78.13{\scriptsize $\pm$ 3.13} & 69.79{\scriptsize $\pm$ 2.39} & 42.22{\scriptsize $\pm$ 3.85} & 65.63{\scriptsize $\pm$ 3.13} & 70.00{\scriptsize $\pm$ 20.00} & 67.67{\scriptsize $\pm$ 1.53} & 2749.94{\scriptsize $\pm$ 27.18} \\
\rowcolor{preprintAbstractBg}
\method{} & 82.29{\scriptsize $\pm$ 3.93} & \textbf{75.00}{\scriptsize $\pm$ \textbf{3.13}} & \textbf{47.78}{\scriptsize $\pm$ \textbf{1.92}} & \textbf{82.29}{\scriptsize $\pm$ \textbf{3.61}} & \textbf{96.67}{\scriptsize $\pm$ \textbf{5.77}} & \textbf{75.50}{\scriptsize $\pm$ \textbf{1.80}} & 2232.65{\scriptsize $\pm$ 32.72} \\
\bottomrule
\end{tabular}
\end{table}

\subsection{Evolution Efficiency (Q2)}
\label{sec:evolution_efficiency}

\begin{wraptable}{r}{0.42\textwidth}
\vspace{-1.56cm}
\caption{Evolution tokens (millions, $\downarrow$) with Qwen3.8-27B on two benchmarks.}
\label{tab:cost}
\centering
\begin{tabularx}{\linewidth}{lcc}
\toprule
Method & BEAM & LongMemEval \\
\midrule
\mstar{} & 76.5{\scriptsize $\pm$ 7.1} & 202.3{\scriptsize $\pm$ 10.3} \\
EvolveMem & 69.2{\scriptsize $\pm$ 7.3} & 102.7{\scriptsize $\pm$ 8.4} \\
\rowcolor{preprintAbstractBg}
\method{} & 73.7{\scriptsize $\pm$ 6.4} & 143.3{\scriptsize $\pm$ 9.6} \\
\bottomrule
\end{tabularx}
\end{wraptable}

We compare the token consumption across self-evolving methods under the same 20-round evolution budget. The results are summarized in \Cref{tab:cost}. LongMemEval incurs higher evolution costs because each question has an independent history, resulting in higher per-round extraction costs than BEAM. Overall, EvolveMem consumes fewer tokens because it evolves only the retrieval and answer configurations, whereas \mstar{} and \method{} evolve the full memory program and incur additional extraction costs. Despite this, \method{} reduces evolution cost by 4\%--30\% over \mstar{}, enabled by compact diagnostic contexts and revision-aware execution reuse introduced in \Cref{sec:implementations}.

\subsection{Ablation Study (Q3)}
\label{sec:ablations}

\begin{wraptable}{R}{0.5\textwidth}
  \vspace{-0.6cm}
\caption{Ablations with Qwen3.8-27B on BEAM-1M. $\Delta$ is relative to full \method{}.}
\label{tab:ablations}
\centering
\setlength{\tabcolsep}{4pt}
\begin{tabularx}{\linewidth}{@{}lrr@{}}
\toprule
\multirow{2}{*}{Variant} & \multicolumn{1}{c}{Overall score} & \multicolumn{1}{c@{}}{$\Delta$} \\
& \multicolumn{1}{c}{(\%) $\uparrow$} & \multicolumn{1}{c@{}}{(pp)} \\
\midrule
\rowcolor{preprintAbstractBg}
Full \method{} & \textbf{63.39}{\scriptsize $\pm$ \textbf{1.02}} & -- \\
w/o capability evolution & 53.24{\scriptsize $\pm$ 2.38} & $-$10.15 \\
w/o trace-guided integration & 60.23{\scriptsize $\pm$ 1.86} & $-$3.16 \\
Round-robin target selection & 60.84{\scriptsize $\pm$ 1.13} & $-$2.55 \\
Random case selection & 58.13{\scriptsize $\pm$ 1.25} & $-$5.26 \\
\bottomrule
\end{tabularx}
\end{wraptable}

To examine the contribution of each design choice, we compare four variants of \method{} on BEAM with Qwen3.8-27B. The results are shown in \cref{tab:ablations}. Removing capability-driven evolution falls back to conventional holistic evolution and reduces the score by 10.15 pp. Without capability-level optimization, mixed feedback can blur optimization directions, while useful capability-specific revisions may not receive further refinement when they do not improve overall performance. This large drop highlights capability-driven evolution as a primary source of improvement.

Removing trace-guided integration reduces the score by 3.16 pp, but the variant still outperforms the baselines. This shows that optimizing along explicit capability directions already improves evolution by reducing mixed feedback and focusing revisions, while integration guided by evaluated programs and paired differential cases across capabilities provides additional gains in overall performance.

Using a fixed round-robin schedule instead of dependency-aware capability selection reduces the score by 2.55 pp, supporting targeted refinement of weak capabilities and potential bottlenecks for multi-capability tasks. Random case selection reduces the score by 5.26 pp, highlighting the importance of matching diagnostic cases to each stage's objective. Our case selection provides capability-relevant failures for focused refinement and contrasting execution results that expose useful revisions and potential side effects during integration.

\subsection{Evolution Trace Analysis (Q4)}
\label{sec:case_analysis}

\begin{figure}[t!]
  \centering
  \captionsetup[subfloat]{font=normalsize}
  \subfloat[Score gains during capability refinement stage\label{fig:evolution_trace_gains}]{%
    \includegraphics[width=0.65\linewidth]{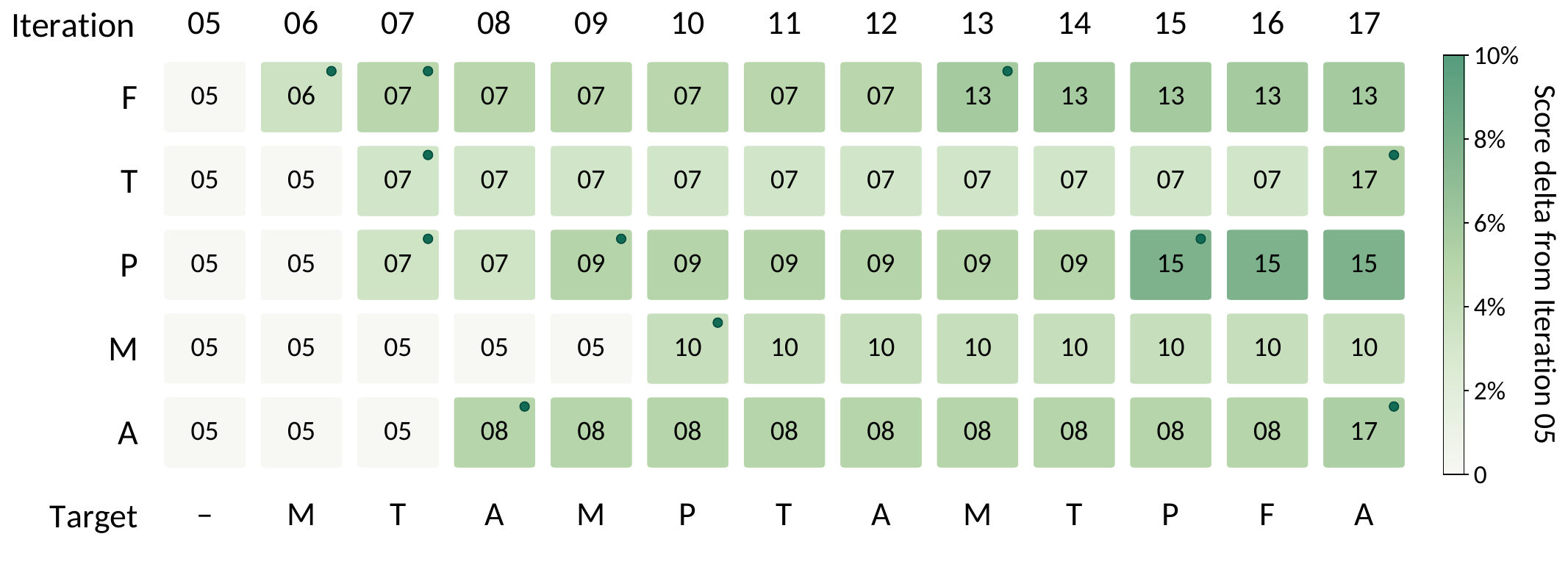}%
  }\hfill
  \subfloat[Peak vs. final performance\label{fig:evolution_trace_peak}]{%
    \includegraphics[width=0.34\linewidth]{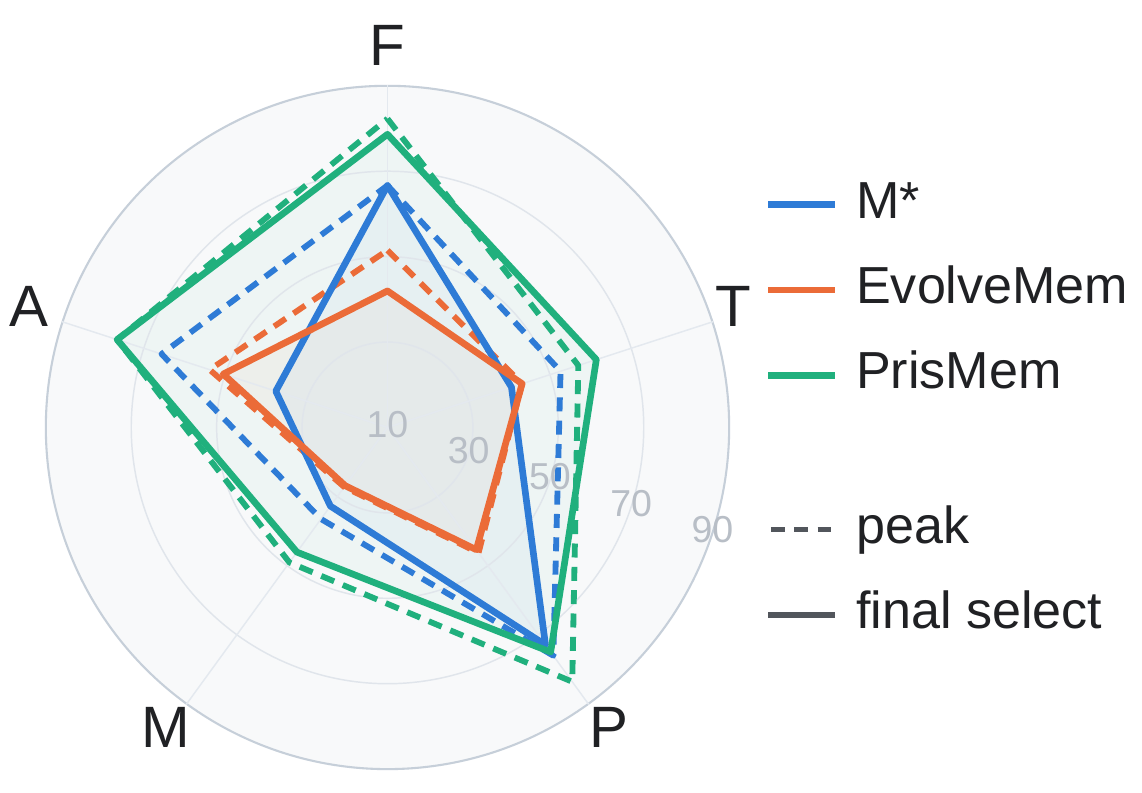}%
  }
  \caption{\textbf{Evolution trace analysis.} 
  (a) Gains of each Stage II iteration relative to program 05 obtained after cold start. Cell numbers are the creation iteration of the retained specialist for each capability; dots mark specialist updates. Scores improve intermittently across capabilities as refinement progresses. 
  (b) Capability-wise scores of \mstar{}, EvolveMem, and \method{}. Solid lines show the final selected programs, while dashed lines show the best capability-wise score reached across all iterations. \method{} outperforms holistic evolution across capabilities, expanding the performance boundary, while its final integrated program closely matches its historical capability-wise best.}
  \label{fig:evolution_trace}
\end{figure}

We analyze \method{}'s evolution trajectories on BEAM with Qwen3.8-27B. In~\Cref{fig:evolution_trace_gains}, capability scores improve intermittently throughout Stage II, showing refinements progressively expand performance along each capability dimension. Refining one capability can also update specialists for others, reflecting dependencies among capabilities and our dependency-aware selection of targets that can benefit multiple capabilities simultaneously. \Cref{fig:evolution_trace_peak} shows that capability-driven evolution expands beyond the performance boundary of holistic evolution, uncovering better optimization directions. Trace-guided integration then successfully consolidates these gains, with the final program closely matching the peak capability-wise performance reached during evolution.

%% file: sections/conclusion.tex
\section{Conclusion}
\label{sec:conclusion}

We introduced capability-driven evolution to expand memory program optimization along explicit capability dimensions.
Capability-specific feedback clarifies revision directions, while capability-level evaluation preserves gains obscured by overall scores.
\method{} combines dependency-aware capability selection and history-guided diagnosis to refine capability specialists, then uses trace-guided integration to consolidate their complementary gains into a unified memory program.
On BEAM and LongMemEval, \method{} outperforms the strongest baselines by 7.83--10.54 pp on million-token histories.
We hope this work encourages further exploration of capability-driven evolution for building stronger agent memory.

%% file: sections/appendix.tex
\tcbset{papercard/.style={
  enhanced,breakable,colback=panelgray,colframe=accent!80!black,
  colbacktitle=accent!80!black,coltitle=white,fonttitle=\bfseries,
  boxrule=0.5pt,arc=2pt,left=6pt,right=6pt,top=5pt,bottom=5pt}}
\newtcolorbox{promptcard}[1]{papercard,title={#1}}
\newcommand{\promptsection}[1]{%
  \par\addvspace{6pt}\noindent\textbf{#1}\ \ignorespaces}

\section{Implementation Details}
\label{sec:implementation_details}

\subsection{Capability Selection and Offline Annotation}
\label{sec:capability_details}

\paragraph{Selecting Capabilities for Our Study.}
Capability-driven evolution needs a vocabulary that separates recurring memory requirements and makes their progress observable.
As an initial step toward capability-driven evolution, we derive five representative capabilities in~\cref{tab:vocabulary} from a broad review of agent memory benchmarks and memory systems~\citep{maharana2024locomo,wu2025longmemeval,tavakoli2026beam,zhong2024memorybank,rasmussen2025zep,chhikara2025mem0,xu2025amem}.
Explicit fact recovery, changing states, persistent user constraints, evidence distributed across sessions, and insufficient or conflicting support recur across these benchmarks and systems.
These requirements provide a concrete instantiation of the vocabulary used in our study, covering \emph{factual retrieval}, \emph{temporal tracking}, \emph{preference extraction}, \emph{multi-session synthesis}, and \emph{adversarial}. 
\Cref{tab:benchmark_capabilities} connects this vocabulary to existing benchmark question types.

\begin{table}[htbp]
\caption{Correspondence between benchmark question types and the memory capabilities used in our study. The entries describe primary requirements; individual questions may involve additional capabilities.}
\label{tab:benchmark_capabilities}
\centering
\begin{minipage}{0.80\linewidth}
\begin{tabularx}{\linewidth}{>{\raggedright\arraybackslash}p{7cm}l}
\toprule
Question type & Primary memory capabilities \\
\midrule
\multicolumn{2}{l}{\textbf{LoCoMo}~\citep{maharana2024locomo}} \\
Single-hop; Open-domain knowledge\textsuperscript{1} & Factual retrieval \\
Multi-hop & Multi-session synthesis \\
Temporal reasoning & Temporal tracking \\
Adversarial & Adversarial \\
\midrule
\multicolumn{2}{l}{\textbf{LongMemEval}~\citep{wu2025longmemeval}} \\
Single-session user; single-session assistant & Factual retrieval \\
Single-session preference & Preference extraction \\
Multi-session & Multi-session synthesis \\
Temporal reasoning; knowledge update & Temporal tracking \\
Abstention variants\textsuperscript{2} & Adversarial \\
\midrule
\multicolumn{2}{l}{\textbf{BEAM}~\citep{tavakoli2026beam}} \\
Information extraction & Factual retrieval \\
Event ordering; temporal reasoning; knowledge update & Temporal tracking \\
Preference following; instruction following & Preference extraction \\
Multi-session reasoning; summarization & Multi-session synthesis \\
Contradiction resolution; Abstention & Adversarial \\
\bottomrule
\end{tabularx}
\par\smallskip
\footnotesize\raggedright
\textsuperscript{1}LoCoMo's open-domain questions require retrieving information about the speaker and combining it with external knowledge. We therefore map this question type to factual retrieval.\par
\textsuperscript{2}The question identifier ending in \texttt{\_abs}.
\end{minipage}
\end{table}

An individual task can require several capabilities.
For example, in BEAM~\citep{tavakoli2026beam}, resolving conflicting statements about whether a user has implemented a project feature requires factual retrieval to recover both claims and adversarial handling to recognize their inconsistency and request clarification.
In LongMemEval~\citep{wu2025longmemeval}, counting the distinct museums or galleries visited in a given month requires multi-session synthesis to combine reported visits and temporal tracking to exclude visits outside that month.
These examples motivate the primary--secondary annotation below.

\paragraph{Offline Capability Annotation.}
\label{sec:capability_annotation}
Question text alone is insufficient for reliably identifying a task's memory requirements.
For example, a BEAM question asks about a user's education and specialization in psychology, yet neither detail is provided in the history.
The task therefore requires adversarial abstention despite its factual wording.
We tested question-only annotation with GPT-5.5 on 700 BEAM-1M questions.
The predicted primary capability matched the benchmark-type mapping in \cref{tab:benchmark_capabilities} for only 337 questions (48.14\%), indicating the limited reliability of question-only annotation in this setting.

We therefore provide the annotator with the question, reference answer and evidence.
The answer clarifies the expected response, while the evidence identifies the information needed to support it.
We use LLM-based annotation so that the procedure also applies when benchmark-specific question types are unavailable.
Given these inputs and the capability definitions, we use GPT-5.5 to assign one primary capability and up to two distinct secondary capabilities using the prompt template below.
The primary label captures the requirement whose failure would most directly prevent a correct answer; secondary labels capture additional requirements.
We reuse valid annotations across runs.

Capability labels are used only to aggregate capability-level feedback and select the next capability to evolve; they are never passed to the memory program.
We therefore annotate only the training and validation sets, with no annotation needed for the test set.
At test time, the evolved program receives the history and question, preserving the same task information available to the baselines for a fair comparison.
Native benchmark types will be used to group the test set for capability-level reporting, as detailed in \cref{sec:dataset_details}.

\begin{promptcard}{Prompt for offline capability annotation}
\promptsection{System.}\par
You are a classifier agent. Classify the question according to the memory-related capabilities it requires. Return exactly the requested JSON object.

\promptsection{Task.}\par
Determine which long-term-memory capabilities are required to answer the current question correctly. Use the question, golden evidence, and golden answer together to identify the memory operation the question tests.

\promptsection{Capabilities.}\par
\begin{itemize}[leftmargin=*]
\item \texttt{factual\_retrieval}: locate and preserve an explicit fact, name, exact value, list item, or other localized detail. Use it as primary when direct information recovery is the central requirement.
\item \texttt{temporal}: order events, compare times, determine recency, or track an old state, a new state, and a knowledge update.
\item \texttt{preference\_extraction}: recover and apply a user's lasting preference, prior instruction, constraint, or standing need.
\item \texttt{multi\_session\_synthesis}: connect, aggregate, or summarize evidence distributed across multiple sessions or events.
\item \texttt{adversarial}: abstain when memory lacks supporting evidence, or cautiously identify and resolve contradictory or misleading evidence.
\end{itemize}
Choose exactly one primary capability: the capability whose failure would most directly prevent a correct answer. Add zero, one, or two distinct secondary capabilities only when they are also genuinely required. For example, a contradiction whose resolution depends on recency can use adversarial as primary and temporal as secondary; a changed preference can use preference extraction and temporal.

\promptsection{Question:}\par
\emph{[question]}

\promptsection{Golden evidence:}\par
\emph{[reference snippets serialized as JSON]}

\promptsection{Golden answer:}\par
\emph{[reference answer]}

\promptsection{Output.}\par
Return exactly one JSON object and no other text. Allowed labels are the five identifiers above; secondary labels must be distinct and different from the primary label.\par
\texttt{\{"primary": "<one allowed capability>",}\par
\texttt{\phantom{\{}"secondary": ["<zero to two other capabilities>"]\}}
\end{promptcard}

\subsection{Evidence-Based Diagnostic Metrics}
\label{sec:diagnostic_details}

During the cold-start stage, we diagnose the memory pipeline by tracing each reference evidence snippet from its source dialogue segment through extraction, indexing, and retrieval. Specifically, we examine whether the relevant information is preserved in the extracted memory units, effectively represented and accessible through the index, and ultimately retrieved for the actual question. These diagnostics allow us to distinguish weaknesses in \textit{Extraction, Indexing, Planning} and \textit{Retrieval} across capabilities.

The three diagnostics use the same embedding-based matching function. For each task $x$, let $E_x$ denote the set of distinct reference evidence snippets annotated by the dataset. Each $e\in E_x$ can be uniquely mapped to its source dialogue segment after text normalization, from which we can obtain the corresponding extracted memory units $U_e$. To measure how well a set of text items preserves or exposes a reference evidence snippet, we compare the evidence with the most similar item in the set. Specifically, for an L2-normalized embedding function $\phi$ and a set of text items $Z$, we define the maximum matching similarity as:
\begin{equation}
\operatorname{sim}_{\max}(e,Z)=
\begin{cases}
\max_{z\in Z}\phi(e)^\top\phi(z), & Z\neq\varnothing,\\
0, & Z=\varnothing.
\end{cases}
\label{eq:match}
\end{equation}
This score measures whether the information in evidence $e$ is represented by at least one item in $Z$. Let $R_e$ denote the text blocks retrieved from the memory index when the original evidence $e$ is used as an oracle query, following the standard query planning and retrieval procedure. Let $R_x$ denote the text blocks retrieved by the memory program for the actual question $x$. We then compute:
\begin{align}
d_{\mathrm{ext}}(x)
&=
\frac{1}{|E_x|}
\sum_{e\in E_x}
\operatorname{sim}_{\max}(e,U_e),
\label{eq:diag_extraction}
\\
d_{\mathrm{idx}}(x)
&=
\frac{1}{|E_x|}
\sum_{e\in E_x}
\operatorname{sim}_{\max}(e,R_e),
\label{eq:diag_index}
\\
d_{\mathrm{ret}}(x)
&=
\frac{1}{|E_x|}
\sum_{e\in E_x}
\operatorname{sim}_{\max}(e,R_x).
\label{eq:diag_retrieval}
\end{align}
Here, $d_{\mathrm{ext}}$ measures whether the relevant evidence is preserved after extraction, $d_{\mathrm{idx}}$ measures whether the preserved information can be effectively accessed through the index using an oracle evidence query, and $d_{\mathrm{ret}}$ measures whether the memory program retrieves the relevant evidence for the actual question. We report these diagnostics separately for each capability. Comparing them helps localize weaknesses to memory extraction, indexing, or query planning and retrieval. For example, a high $d_{\mathrm{ext}}$ but substantially lower $d_{\mathrm{idx}}$ suggests that the evidence is preserved in memory units but not effectively indexed or accessed. A high $d_{\mathrm{idx}}$ but low $d_{\mathrm{ret}}$ instead points to problems in query formulation, query planning, or retrieval.

Two length statistics provide additional context: the average number of characters of the memory units and retrieved content. The former helps assess whether extraction produces overly verbose memory units, while the latter characterizes the amount of information exposed to the answer model and provides context for interpreting retrieval quality.

\subsection{Diagnostic Case Presentation and Compression}
\label{sec:case_details}

Diagnostic cases provide the context needed to understand and diagnose errors in the memory pipeline, especially in later evolution iterations. To support detailed analysis, each case presents a comprehensive view of the information relevant to the task. Each case includes the primary and secondary capability labels, question, reference answer, predicted answer, task score, query plan, retrieved text, and a reference evidence view. The evidence view pairs each reference evidence snippet with its original source text and the memory units in which it is represented. This pairing helps distinguish missing evidence from evidence that was preserved in memory but poorly indexed, retrieved, or interpreted. Diagnosis and coding receive the same compact case material, allowing each proposed change to be traced back to the evidence that motivates it.

Full source segments and retrieved passages can be lengthy, making diagnostic cases unnecessarily large. We therefore compress the material presented to the evolution model while preserving the evidence needed for diagnosis. We use a budget of 2400 characters per displayed excerpt and retain shorter texts in full. 
For longer texts, we divide them into source records, paragraphs, or sentences, while preserving session, date, and role headers and keeping fenced code blocks intact. A greedy selection procedure then selects spans that maximize coverage of the question and reference answer, corresponding reference evidence, and the model prediction. Span relevance combines semantic cosine similarity and stemmed non-stopword Dice overlap, weighted at 0.75 and 0.25, respectively. The question--answer cue has weight 1, the evidence snippets have a total weight of 1, and the prediction has weight 0.5. Including the prediction helps retain evidence that explains an incorrect answer, such as conflicting numeric values.

Selected spans are restored to their original source order and separated by explicit omission markers (\texttt{...}). For extracted memory, we always retain at least the best-matching unit for each reference evidence snippet, add other relevant units when space permits, and deduplicate repeated displays. These choices preserve the connections among reference evidence, its memory representation, and the model prediction while removing repeated or unrelated context. We evaluate the effect of context reduction on \method{}'s real traces from BEAM and find that the average case-context size decreases from 16,776 to 9,428 tokens, corresponding to a 43.8\% reduction.

\subsection{Validate Generated Programs}
\label{sec:program_checks}

Generated memory program revisions can fail before they can be evaluated.
We check program completeness, interface compatibility, and basic execution behavior before allocating task model calls, so readily detectable errors can be corrected early.
The checks proceed from inexpensive source inspection to bounded synthetic execution.

\begin{enumerate}[leftmargin=*]
\item \textbf{Source and module completeness.}
The patched source must be nonempty, parse as Python, use allowed imports, and avoid evaluation-side interfaces.
The module must load successfully and define \texttt{MemoryUnit}, \texttt{Query}, \texttt{Memory}, and the three nonempty extraction, planning, and answer instruction strings.
A patch must make an actual source change.
\item \textbf{Schema and signature compatibility.}
\texttt{MemoryUnit} and \texttt{Query} must be nonempty dataclasses with supported JSON-compatible fields: strings, integers, floats, booleans, string lists, and their optional forms.
The constructor, \texttt{index}, and \texttt{retrieve} must preserve the required synchronous signatures.
Structured outputs are checked for missing or extra fields, invalid types, and nonfinite numeric values.
\item \textbf{Functional smoke tests.}
A local stub backend exercises construction, indexing a session with synthetic units, indexing an empty-unit session, and retrieval with a valid query.
Indexing must return \texttt{None}; retrieval must return a string within the 16K-character context limit.
These tests require no task-LLM inference.
\item \textbf{Synthetic complexity checks.}
A larger probe indexes $\sim$2K synthetic units across 100 batches and executes two queries with 4096-dimensional embeddings.
It checks local computation and backend-call counts.
A local CPU-work budget and an outer process timeout bound stalled or unexpectedly expensive execution.
\end{enumerate}

These checks catch malformed schemas, incomplete patches, interface mismatches, and basic execution failures before full task evaluation.

\subsection{Memory Interface.}
\label{sec:memory_interface}

The memory interface defines the scope of modifications during self-evolution. We decompose the memory program into five components: \emph{Extraction}, \emph{Indexing}, \emph{Planning}, \emph{Retrieval}, and \emph{Answer}, enabling precise problem localization and targeted modifications by the model.

The five components are defined as follows: \emph{Extraction} converts history segments into structured memory units $U=P.\mathrm{extract}(\mathcal{H})$; \emph{Indexing} organizes these units $U$ into persistent memory $M=P.\mathrm{index}(U,\mathcal{H})$; \emph{Planning} translates the question $q$ into a structured query unit $Q=P.\mathrm{plan}(q)$; \emph{Retrieval} executes the query $Q$ against persistent memory $M$ to return relevant content $R=P.\mathrm{retrieve}(M,Q)$; and \emph{Answer} instructs the task model to use the retrieved content $R$ and question $q$ to generate a high-quality answer $\hat{y}=P.\mathrm{answer}(q,R)$.
Evolution can modify the schemas of memory and query units, as well as the code and instructions for these five memory components.

Here are the memory interface definitions provided in the prompt.

\begingroup
\lstset{basicstyle=\ttfamily,breaklines=true,breakatwhitespace=false,
  columns=fullflexible,keepspaces=true,showstringspaces=false,tabsize=4}
\begin{promptcard}{Memory interface definition}
You are designing a Memory Program that implements three classes:

\promptsection{1. MemoryUnit (dataclass):} Defines what information is captured as memory units from source text.
\begin{itemize}[leftmargin=*]
\item Must be a \texttt{@dataclass} with typed fields
\item An external LLM will populate several instances by generating JSON matching your field definitions
\item \textbf{Field types MUST be JSON-compatible}: use only \texttt{str}, \texttt{int}, \texttt{float}, \texttt{bool}, \texttt{list[str]}, and their Optional forms
\item Do NOT use datetime, tuple, bytes, dict, nested dataclasses, or custom objects --- the fixed harness cannot construct them from the generated JSON
\item Use \texttt{field(metadata=\{"description": "..."\})} to describe fields --- descriptions are shown to the LLM that populates instances
\end{itemize}

\promptsection{2. Query (dataclass):} Defines what parameters are used when retrieving from memory.
\begin{itemize}[leftmargin=*]
\item Must be a \texttt{@dataclass} with typed fields
\item An external LLM will populate one instance by generating JSON matching your field definitions
\item Same JSON-compatible type restriction and field description support as MemoryUnit
\end{itemize}

\promptsection{3. Memory (class):} The core memory system.
\begin{lstlisting}
__init__(self, backend)
\end{lstlisting}
Receives a Backend with:
\begin{lstlisting}
backend.ask(prompt, system_prompt=None,
            json_object=False) -> str
\end{lstlisting}
LLM for reasoning, summarization, or reranking
\begin{lstlisting}
backend.embed(texts: list[str]) -> np.ndarray
\end{lstlisting}
One L2-normalized embedding row per input text
\begin{lstlisting}
index(self, units: list[MemoryUnit], raw_text: str) -> None
\end{lstlisting}
Incrementally index one source session. \texttt{units} contains all valid MemoryUnits generated from that session and may be empty. \texttt{raw\_text} is the raw session text with a timestamp. Calls occur in chronological order. In this function, you may merge true duplicates, normalize references to the same entity, preserve timestamps and the order of distinct, updated, or conflicting facts, and build dense, lexical, exact-value, graph, metadata, or combined indexes for retrieval.
\begin{lstlisting}
retrieve(self, query: Query) -> str
\end{lstlisting}
Retrieve relevant information as a string. Use the Query fields to search the relevant dense, lexical, exact-value, graph, or metadata indexes, rank and combine matches, remove repeated results, and return the useful context within \texttt{OUT\_CHAR\_BUDGET}.

Allowed imports: json, re, math, hashlib, collections, dataclasses, typing, datetime, textwrap, numpy. The execution environment provides \texttt{OUT\_CHAR\_BUDGET}, the enforced retrieval-character limit.

\promptsection{Runtime constraints}

These limits are enforced during evaluation. Violating them makes the evaluation fail.
\begin{itemize}[leftmargin=*]
\item \textbf{\texttt{retrieve()} output limit}: \texttt{Memory.retrieve()} must return at most \textbf{16000 characters}.
\item \texttt{retrieve()} is called once per question, and must be read-only and independent of prior query calls.
\end{itemize}

\promptsection{Instruction constants (required)}

Three module-level string constants provide the natural-language instructions used in task-LLM prompts:
\begin{itemize}[leftmargin=*]
\item \texttt{MemoryExtractionInstruction}: Instruction for MemoryUnit generation. Tells the task LLM to extract atomic facts, procedures, relationships, decisions, and preferences from each segment, resolve references, and preserve names, exact values, dates/times, and lists. The MemoryUnit schema is provided separately --- no need to describe field names or types here.
\item \texttt{QueryPlanningInstruction}: Instruction for Query generation. Tells the task LLM how to formulate retrieval queries and tailor them to the capabilities required by the question, including splitting multi-aspect questions when useful. The Query schema is provided separately --- no need to describe field names or types here.
\item \texttt{AnswerInstruction}: Instruction for answer generation. Controls answer format, length, and style, and guides how retrieved content is interpreted and used. Apply question-specific strategies when needed, including distinguishing user statements from assistant suggestions, resolving updates or contradictions using source, time, and scope, counting distinct events rather than memory units or repeated mentions, and synthesizing facts across sessions. Preserve relevant chronology and specific changes for progression questions. Ensure that information needed for source, time, or event identification is preserved in memory and exposed during retrieval.
\end{itemize}

MemoryExtractionInstruction, QueryPlanningInstruction, and AnswerInstruction must not be empty.
\end{promptcard}
\endgroup

\subsection{Revision-Aware Execution Reuse}
\label{sec:reuse_details}

A local revision often leaves most of the memory pipeline unchanged.
We compare the generated candidate's components with those of its parent and begin re-execution at the earliest affected stage, as shown in~\cref{tab:reuse}.
For example, a new ranking rule can reuse extracted units and the built memory, whereas a changed extraction instruction requires new units and all downstream computations.
This dependency-aware reuse avoids unnecessary LLM calls while preserving evaluation coverage.

\begin{table}[htbp]
\caption{Re-execution boundaries for a candidate relative to its parent. For changes spanning several components, the earliest affected stage determines what can be reused.}
\label{tab:reuse}
\centering
\renewcommand{\tabularxcolumn}[1]{m{#1}}
\begin{tabularx}{\linewidth}{>{\raggedright\arraybackslash}m{4.5cm}>{\raggedright\arraybackslash}m{3.5cm}Y}
\toprule
Changed component & Re-execution begins at & Reusable work \\
\midrule
Memory-unit schema or extraction stage & Extraction & Segmentation \\
\addlinespace[3pt]
Constructor, index, or helpers & Indexing & Segmentation and extracted units \\
\addlinespace[3pt]
Query schema, planning, or retrieval & Query planning & Segmentation, extraction, and indexed memory \\
\addlinespace[3pt]
Answer instruction only & Answer generation & Built memory, query plan, and retrieved context \\
\bottomrule
\end{tabularx}
\end{table}

Reusable state is copied before being attached to the candidate, keeping parent and child memory states separate.
Changes to shared helpers invalidate indexing conservatively because those helpers may affect memory construction.
Reuse follows the program's execution dependencies and the actual inputs to cached computations.

\subsection{Seed Memory Program}
\label{sec:seed_program}

The initial seed memory program is deliberately simple, providing a broad search space. It appends extracted units in input order and returns their concatenation, truncated to the context budget, without query-dependent ranking. The execution environment fixes the output character budget at 16K (\texttt{OUT\_CHAR\_BUDGET = 16000}) across evolution runs. The implementation is shown below.

\tcbinputlisting{papercard,colframe=orange!70!black,colbacktitle=orange!70!black,
  title={Initial seed memory program used for \method{}},
  listing only,listing file={programs/seed.py},
  listing options={language=Python,basicstyle=\ttfamily,breaklines=true,
    breakatwhitespace=false,columns=fullflexible,keepspaces=true,
    showstringspaces=false,tabsize=4}}

\subsection{Pseudocode of \method{}}

We provide the pseudocode of \method{} below, illustrating its execution flow.

\begin{algorithm}[htbp]
\caption{\method{}: Capability-Driven Memory Evolution}
\label{alg:evolution}

\begin{algorithmic}[1]
\Require Seed $P_0$; datasets $\mathcal{S}_{\mathrm{train}},\mathcal{S}_{\mathrm{valid}}$; capability set $\caps$; iterations for three stages $T_1, T_2, T_3$ with $T=T_1+T_2+T_3$; case budget $K$; thresholds $\delta, \theta$.
\Ensure Final memory program $G_T$.
\State $\mathcal{W}\gets\{P_0\}$; \Call{Evaluate}{$P_0,\mathcal{S}_{\mathrm{train}},\mathcal{S}_{\mathrm{valid}}$}
\For{$t=1,\ldots,T_1$} \Comment{\textbf{Stage I: Cold Start}}
  \State $P\gets\Call{Select}{\mathcal{W},t}$; $M\gets\Call{ComputeMetrics}{P, \mathcal{S}_{\mathrm{train}}}$
  \State $d\gets\Call{Diagnose}{P,M}$; $P'\gets\Call{Revise}{P,d,M}$
  \If{$P'\neq\bot$ and \Call{Evaluate}{$P',\mathcal{S}_{\mathrm{train}},\mathcal{S}_{\mathrm{valid}}$} succeeds}
    \State $\mathcal{W}\gets\mathcal{W}\cup\{P'\}$
  \EndIf
\EndFor
\State $P_{\mathrm{base}}\gets\argmax_{P\in\mathcal{W}}s(P;\mathcal{S}_{\mathrm{valid}})$; $\pool_{T_1}\gets\{P_{\mathrm{base}}\}$; $G_{T_1}\gets P_{\mathrm{base}}$
\State $P_{T_1}^\star(c)\gets P_{\mathrm{base}}$ for every $c\in\caps$

\For{$t=T_1+1,\ldots,T_1+T_2$} \Comment{\textbf{Stage II: Capability Refinement}}
  \State $c_t\gets\Call{SelectCapability}{\{P_{t-1}^\star(c)\}_{c\in\caps},t}$ \Comment{Priority $\rho_t(c)$ via \cref{eq:target_priority}}
  \State $P_t\gets P_{t-1}^\star(c_t)$
  \State $\mathcal{B}_t\gets\Call{TopK}{\{x_i\in\mathcal{S}_{\mathrm{train}}^{c_t}:s(P_t;x_i)<\delta\}, w_t, \delta, K}$ \Comment{Curate via \cref{eq:case_weight}}
  \State $\mathcal{Z}_t\gets\{\Call{Case}{P_t,x}\mid x\in\mathcal{B}_t\}$
  \State $d_t\gets\Call{Diagnose}{P_t,c_t,\mathcal{Z}_t}$
  \State $P'\gets\Call{Revise}{P_t,d_t,\mathcal{Z}_t}$
  \If{$P'\neq\bot$ and $s(P';\mathcal{S}_{\mathrm{train}}^{c_t})>s(P_t;\mathcal{S}_{\mathrm{train}}^{c_t})$}
    \If{\Call{Evaluate}{$P',\mathcal{S}_{\mathrm{train}},\mathcal{S}_{\mathrm{valid}}$} succeeds} \Comment{Full-suite evaluation}
      \State $\pool_t\gets\pool_{t-1}\cup\{P'\}$
    \Else
      \State $\pool_t\gets\pool_{t-1}$
    \EndIf
  \Else
    \State $\pool_t\gets\pool_{t-1}$
  \EndIf
  \State $P_t^\star(c) \gets \argmax_{P\in\pool_t}s(P;\mathcal{S}_{\mathrm{valid}}^{c})$ for every $c\in\caps$
  \State $G_t\gets\argmax_{P\in\pool_t}s(P)$
\EndFor

\State $\mathcal{E}\gets\varnothing$; $\mathcal{B}\gets\varnothing$; $T' \gets T_1+T_2$ \Comment{\textbf{Stage III: Integration}}
\State $G_T \gets G_{T'}$; $\mathcal{T}\gets\Call{EvolutionTrace}{\pool_{T'}}$; 
\ForAll{$c\in\caps$ where $s(P_{T'}^\star(c);\mathcal{S}_{\mathrm{train}}^c) > s(G_{T'};\mathcal{S}_{\mathrm{train}}^c)$}
  \State $\hat{x}_c\gets\argmax_{x\in\mathcal{S}_{\mathrm{train}}^c}\bigl[s(P_{T'}^\star(c);x)-s(G_{T'};x)\bigr]$
  \State $\mathcal{E}\gets\mathcal{E}\cup\{\Call{Case}{P_{T'}^\star(c),\hat{x}_c}, \Call{Case}{G_{T'},\hat{x}_c}\}$
  \State $\mathcal{B}\gets\mathcal{B}\cup\{\hat{x}_c\}$
\EndFor
\If{$\mathcal{E}\neq\varnothing$}
  \State $\Pi\gets\Call{PlanIntegration}{G_{T'},\{P_{T'}^\star(c)\}_{c\in\caps},\mathcal{T},\mathcal{E}}$ \Comment{Generate plan}
  \For{$t=T'+1,\ldots,T$}
  \State $P_t\gets\Call{ExecuteIntegrate}{G_{T'},\Pi,\mathcal{E}}$ \Comment{Coding with plan and paired cases}
  \If{$P_t\neq\bot$ and $s(P_t)>s(G_{T'})$ and $\max_{c\in\caps}[1 - s(P_t;\mathcal{S}_{\mathrm{train}}^c) / s(G_{T'};\mathcal{S}_{\mathrm{train}}^c)]\le \theta$}
    \State $G_T\gets P_t$
    \State \textbf{break}
  \Else 
  \State $\mathcal{E} \gets \mathcal{E} \cup\{\Call{Case}{P_t,x} \mid x \in \mathcal{B}\}$ \Comment{Accumulate cases}
  \EndIf
  \EndFor
\EndIf
\State \Return $G_T$ \Comment{Final unified program}
\end{algorithmic}
\end{algorithm}

\section{Experimental Details}
\label{sec:experiment_details}

\subsection{Dataset Splits and Capability Composition}
\label{sec:dataset_details}

We evolve memory programs on the shorter-history BEAM-100K and LongMemEval-S splits ($\sim$100K tokens) and evaluate their frozen implementations on the million-token BEAM-1M and LongMemEval-M splits. This setup reduces the risk of overestimating evolutionary performance due to overfitting and provides a more direct evaluation of the evolution method's ability to generalize to million-token histories.

For BEAM, the evolution split contains 160 training questions from 8 conversations and 120 validation questions from 6 conversations, while the test set contains all 700 BEAM-1M questions from another 35 conversations.
For LongMemEval, each question is associated with a specific conversation. To reduce evaluation cost while maintaining a balanced question-type distribution, we stratify by question type and randomly sample 60 questions for training and 60 for validation in total. Following the same stratified sampling procedure, we construct a 200-question subset of LongMemEval-M with different histories for testing.
The frozen program constructs memory from each test history with its implementation held fixed.

\Cref{tab:splits} reports the capability counts and average history length of each split.
Training and validation labels are assigned by the LLM annotator in \cref{sec:capability_details}.
Test questions are grouped using the native question-type mapping in \cref{tab:benchmark_capabilities}.

\begin{table}[htbp]
\caption{Sizes, primary capability composition, and average history length of each dataset split. F, T, P, M, and A denote factual retrieval, temporal tracking, preference extraction, multi-session synthesis, and adversarial, respectively. Average history length is measured with the Qwen3.8-27B tokenizer.}
\label{tab:splits}
\centering
\begin{tabularx}{0.88\linewidth}{llrrrrrrr}
\toprule
Dataset & Split & F & T & P & M & A & Total & Avg. tokens (K) \\
\midrule
BEAM-100K & Train & 20 & 43 & 32 & 33 & 32 & 160 & 125.11 \\
BEAM-100K & Validation & 18 & 32 & 24 & 22 & 24 & 120 & 146.83 \\
BEAM-1M & Test & 70 & 210 & 140 & 140 & 140 & 700 & 1,089.57 \\
\midrule
LongMemEval-S & Train & 19 & 17 & 11 & 9 & 4 & 60 & 109.40 \\
LongMemEval-S & Validation & 20 & 17 & 8 & 8 & 7 & 60 & 109.40 \\
LongMemEval-M & Test & 64 & 64 & 30 & 32 & 10 & 200 & 1,079.04 \\
\bottomrule
\end{tabularx}
\end{table}

\subsection{Parameter Settings}
\label{sec:search_details}

\paragraph{Model Inference Parameters.}
For task models, we disable thinking or use low thinking effort with a maximum of 8,192 tokens. For diagnosis and coding models, we enable thinking with medium reasoning effort and a maximum of 65,536 tokens. All methods use the same inference settings. For benchmark judging, we use the same model as the task model and set the temperature to 0 for deterministic outputs. We test Qwen3.8-27B and GPT-5.5 in our experiments.

\begin{table}[htbp]
\caption{Parameter settings of \method{} we used in experiments.}
\label{tab:settings}
\centering
\begin{tabularx}{0.60\linewidth}{Yr}
\toprule
Parameter & Value \\
\midrule
\multicolumn{2}{l}{\textbf{Stage II: Capability refinement}} \\
Case score threshold $\delta$ & 0.8 \\
Refinement case budget $K$ & 5 \\
\midrule
\multicolumn{2}{l}{\textbf{Stage III: Integration}} \\
Capability-wise score degradation ratio $\theta$ & 0.1 \\
Integration case budget & 1 per capability \\
\bottomrule
\end{tabularx}
\end{table}

\paragraph{Algorithm Parameters.}
We use an evolution budget of $T=20$ rounds for all evolutionary methods. For \method{}, we allocate $T_1=5$ rounds to cold start, $T_2=12$ to capability refinement, and $T_3=3$ to final integration.
\Cref{tab:settings} lists the parameter values used for stages II and III.

\section{Additional Experimental Results}

\subsection{Results on GPT-5.5}
\label{sec:gpt_results}

\Cref{tab:gpt_beam} shows the performance of different methods on BEAM using GPT-5.5. The results preserve the same trends observed with Qwen3.8-27B (\cref{tab:main}), demonstrating the robustness of \method{} across different underlying models.

\begin{table}[htbp]
\caption{Performance on BEAM with GPT-5.5.}
\label{tab:gpt_beam}
\centering
\begin{tabularx}{0.5\linewidth}{Ycc}
\toprule
\multirow{2}{*}{Method} & Overall Score & Avg. \#Tokens \\
& (\%) $\uparrow$ & (K) $\downarrow$ \\
\midrule
Mem0 & 57.77{\scriptsize $\pm$ 0.17} & 180.21{\scriptsize $\pm$ 0.12} \\
A-MEM & 60.92{\scriptsize $\pm$ 0.41} & 352.11{\scriptsize $\pm$ 0.25} \\
HippoRAG2 & 57.73{\scriptsize $\pm$ 0.50} & 193.52{\scriptsize $\pm$ 0.18} \\
SimpleMem & 52.36{\scriptsize $\pm$ 0.55} & 121.39{\scriptsize $\pm$ 0.69} \\
LightMem & 46.68{\scriptsize $\pm$ 0.05} & \textbf{94.05}{\scriptsize $\pm$ \textbf{0.19}} \\
\midrule
\mstar{} & 60.75{\scriptsize $\pm$ 2.53} & 97.47{\scriptsize $\pm$ 4.75} \\
EvolveMem & 46.77{\scriptsize $\pm$ 1.56} & 94.39{\scriptsize $\pm$ 3.08} \\
\rowcolor{preprintAbstractBg}
\method{} & \textbf{67.09}{\scriptsize $\pm$ \textbf{1.79}} & 99.36{\scriptsize $\pm$ 3.62} \\
\bottomrule
\end{tabularx}
\end{table}

\subsection{Transfer Experiments}
\label{sec:transfer_experiments}

We examine whether \method{}'s evolved memory programs transfer beyond the model and dataset used during evolution. We evaluate memory programs evolved with Qwen3.8-27B on BEAM-100K. For model transfer, we evaluate these programs with GPT-5.5 on BEAM-1M. For dataset transfer, we retain Qwen3.8-27B and evaluate on LongMemEval-M. The other settings remain the same as those in the main experiments. The results are summarized in \cref{tab:model_transfer} and \cref{tab:dataset_transfer}.

\begin{table}[htbp]
\caption{Model transfer evaluated with GPT-5.5 on BEAM-1M. All programs are evolved on BEAM-100K. Overall scores (\%) are reported as ``mean $\pm$ std'' over three runs.}
\label{tab:model_transfer}
\centering
\begin{tabularx}{0.7\linewidth}{Ycc}
\toprule
Method & \shortstack{Evolved with Qwen3.8-27B\\(transfer)} & \shortstack{Evolved with GPT-5.5\\(target model)} \\
\midrule
\method{} & 67.25{\scriptsize $\pm$ 1.35} & 67.09{\scriptsize $\pm$ 1.79} \\
\bottomrule
\end{tabularx}
\end{table}

\begin{table}[htbp]
\caption{Dataset transfer evaluated with Qwen3.8-27B on LongMemEval-M. All programs are evolved with Qwen3.8-27B. Overall scores (\%) are reported as ``mean $\pm$ std'' over three runs.}
\label{tab:dataset_transfer}
\centering
\begin{tabularx}{0.7\linewidth}{Ycc}
\toprule
Method & \shortstack{Evolved on BEAM\\(transfer)} & \shortstack{Evolved on LongMemEval\\(target dataset)} \\
\midrule
\method{} & 76.50{\scriptsize $\pm$ 1.67} & 75.50{\scriptsize $\pm$ 1.80} \\
\bottomrule
\end{tabularx}
\end{table}

These results show that the evolved memory programs maintain strong performance even when transferred to different models and datasets without additional program optimization, demonstrating the robustness and generalizability of the capability-driven evolution approach.

\subsection{Capability Router Experiments}
\label{sec:capability_router}

We compare trace-guided integration with routing among capability-best programs from the same \method{} evolution trajectory on BEAM. The \emph{question-only router} uses Qwen3.8-27B to predict the primary capability from the question alone, since reference answers and evidence are unavailable at test time. The \emph{oracle-label router} uses the benchmark question types mapped to capabilities in \cref{tab:benchmark_capabilities}. Both routers select the corresponding capability-best program from the evolution run. In contrast, integration uses a single program for all questions without capability labels at test time. Overall scores are computed over all 700 questions in BEAM-1M.

\begin{table}[H]
\caption{Routing versus integration on BEAM-1M with Qwen3.8-27B.}
\label{tab:capability_router}
\centering
\begin{tabularx}{0.42\linewidth}{Yr}
\toprule
Strategy & Overall Score (\%) \\
\midrule
Question-only router & 57.74{\scriptsize $\pm$ 1.21} \\
Oracle-label router & 62.27{\scriptsize $\pm$ 1.14} \\
\rowcolor{preprintAbstractBg}
\method{} & \textbf{63.39}{\scriptsize $\pm$ \textbf{1.02}} \\
\bottomrule
\end{tabularx}
\end{table}

\Cref{tab:capability_router} shows that integration outperforms question-only routing by 5.65 pp and oracle-label routing by 1.12 pp. Question-only routing is limited by the difficulty of inferring memory requirements from the question alone during test time, without access to the reference answer and supporting evidence; its predicted labels match the oracle mapping for only 48\% of the total 700 questions. Although oracle-label routing uses accurate capability labels, it still selects individual programs optimized primarily along specific capability directions. Since a question may require multiple capabilities, these programs may provide unbalanced coverage of its requirements. In contrast, trace-guided integration combines complementary mechanisms within a shared program and reconciles their conditions of use, resulting in stronger overall performance. These results support the value of integrating useful capability-specific revisions into a shared memory program.

\section{Detailed Evolution Trace}
\label{sec:evolution_trace}

\Cref{tab:milestones} summarizes key evolution iterations of \method{} evolved on BEAM with Qwen3.8-27B, including the diagnosed failure, resulting revision, and modified memory program components. Blue tags denote the capability targeted by each evolution step, while gray letter boxes mark capabilities for which the resulting node achieves the best score (specialist).

\newcolumntype{L}[1]{%
  >{\raggedright\arraybackslash}p{#1}%
}
\begingroup
\setlength{\LTcapwidth}{\textwidth}
\begin{longtable}{
@{}
L{0.125\textwidth}
L{0.35\textwidth}
L{0.355\textwidth}
L{\dimexpr0.17\textwidth-6\tabcolsep\relax}
@{}
}
\caption{BEAM evolution traces with failure diagnoses, revision summaries, and current specialist. \textbf{F}: factual retrieval, \textbf{T}: temporal tracking, \textbf{P}: preference extraction, \textbf{M}: multi-session synthesis, \textbf{A}: adversarial.}
\label{tab:milestones}\\

\toprule
Evolution Trace
& Failure diagnosis
& Revision summary
& Current best for\\
\midrule
\endfirsthead

\toprule
Evolution Trace
& Failure diagnosis
& Revision summary
& Current best for\\
\midrule
\endhead

\bottomrule
\endfoot

% =========================================================
% Stage I
% =========================================================

\multicolumn{4}{@{}l}{\textbf{Stage I: Cold-start revisions}} \\

\topcell{%
  \texttt{n05}%
}
&
\topcell{%
  A single flat query could not capture multiple semantic aspects of synthesis questions, causing retrieval to over-concentrate on one topic while missing complementary evidence scattered across sessions.%
}
&
\topcell{%
  Decompose multi-aspect questions into focused sub-queries and retrieve each aspect independently with proportional budget allocation and cross-query deduplication.

  \componentlabels{%
    \schema{}\componentsep%
    \plan{}\componentsep%
    \retrievec{}\componentsep%
    \answer{}%
  }%
}
&
\topcell{%
  \eliteM\elitegap
  \eliteT\elitegap \\
  \eliteA\elitegap
  \eliteP\elitegap \\
  \eliteF
}
\\

% =========================================================
% Stage II
% =========================================================

\midrule
\multicolumn{4}{@{}l}{\textbf{Stage II: Capability-directed refinement}} \\

\topcell{%
  \texttt{n05}$\to$\texttt{n06}\\[2.5pt]
  \targettag{Synthesis}%
}
&
\topcell{%
  Over-atomized memory units fragmented each logical item into multiple small facts, causing related evidence to appear as an unstructured list and hindering distinct-item counting and cross-session progression synthesis.%
}
&
\topcell{%
  Add a topic label to each memory unit and group retrieved evidence by topic, so related facts are presented as coherent blocks for counting and multi-session synthesis.

  \componentlabels{%
    \schema{}\componentsep%
    \extract{}\componentsep%
    \indexc{}\componentsep%
    \retrievec{}\componentsep%
    \answer{}%
  }%
}
&
\topcell{%
  \eliteF
}
\\

\addlinespace[3pt]

\topcell{%
  \texttt{n05}$\to$\texttt{n07}\\[2.5pt]
  \targettag{Temporal}%
}
&
\topcell{%
  Session-level temporal provenance was lost during indexing, so retrieved memories lacked discussion-time context and the model could not reliably distinguish statement order from event dates.%
}
&
\topcell{%
  Add session date to each memory unit, recover it from the source session during indexing, and expose explicit date markers in retrieved evidence while preserving relevance-based ranking.

  \componentlabels{%
    \schema{}\componentsep%
    \extract{}\componentsep%
    \indexc{}\componentsep%
    \retrievec{}\componentsep%
    \answer{}\componentsep%
    \helpers{}%
  }%
}
&
\topcell{%
  \eliteF\elitegap
  \eliteT\elitegap \\
  \eliteP
}
\\

\addlinespace[3pt]

\topcell{%
  \texttt{n05}$\to$\texttt{n08}\\[2.5pt]
  \targettag{Adversarial}%
}
&
\topcell{%
  Memory units lacked speaker provenance, causing user-confirmed facts to be conflated with assistant recommendations or hypotheticals and obscuring genuine contradictions between user statements.%
}
&
\topcell{%
  Add explicit speaker labels to memory units and propagate them through retrieval, then constrain answering to treat user-sourced units as direct evidence and surface conflicting user statements without resolution.

  \componentlabels{%
    \schema{}\componentsep%
    \extract{}\componentsep%
    \indexc{}\componentsep%
    \retrievec{}\componentsep%
    \answer{}%
  }%
}
&
\topcell{%
  \eliteA
}
\\

\addlinespace[3pt]

\topcell{%
  \texttt{n05}$\to$\texttt{n09}\\[2.5pt]
  \targettag{Synthesis}%
}
&
\topcell{%
  Per-unit relevance ranking favored locally similar facts while dropping weaker but essential progression evidence, and the remaining units were returned as a flat list, preventing reconstruction of coherent cross-session threads.%
}
&
\topcell{%
  Add topic-level thread labels, group memories by topic and session order, retrieve matched threads as chronological blocks, and use relevance ranking only to fill remaining uncovered evidence.

  \componentlabels{%
    \schema{}\componentsep%
    \extract{}\componentsep%
    \indexc{}\componentsep%
    \retrievec{}\componentsep%
    \answer{}\componentsep%
    \helpers{}%
  }%
}
&
\topcell{%
  \eliteP
}
\\

\addlinespace[3pt]

\topcell{%
  \texttt{n09}$\to$\texttt{n10}\\[2.5pt]
  \targettag{Preference}%
}
&
\topcell{%
  Standing preferences were retrieved only through topical relevance, so cross-cutting user constraints were often omitted when the current query concerned a different domain.%
}
&
\topcell{%
  Add an explicit standing-preference flag and reserve retrieval capacity for these units, ensuring persistent user constraints are always exposed alongside task-specific evidence.

  \componentlabels{%
    \schema{}\componentsep%
    \extract{}\componentsep%
    \indexc{}\componentsep%
    \retrievec{}\componentsep%
    \answer{}\componentsep%
    \helpers{}%
  }%
}
&
\topcell{%
  \eliteM
}
\\

\addlinespace[3pt]

\topcell{%
  \texttt{n07}$\to$\texttt{n13}\\[2.5pt]
  \targettag{Synthesis}%
}
&
\topcell{%
  The AnswerInstruction lacked explicit synthesis rules for organizing retrieved evidence across topics and time, leading to fragmented narratives and inconsistent handling of counts, cumulative values, and distinct focus areas.%
}
&
\topcell{%
  Add concise answer-side guidance to build chronological topic narratives, distinguish duplicates from cumulative or additive values, and enumerate distinct areas consistently.

  \componentlabels{%
    \answer{}%
  }%
}
&
\topcell{%
  \eliteF
}
\\

\addlinespace[3pt]

\topcell{%
  \texttt{n07}$\to$\texttt{n15}\\[2.5pt]
  \targettag{Preference}%
}
&
\topcell{%
  Retrieved user preferences were treated as passive context rather than binding response constraints, causing the answer LLM to ignore required formatting and presentation behaviors.%
}
&
\topcell{%
  Strengthen the AnswerInstruction to enforce retrieved standing preferences as mandatory constraints on response structure and content.

  \componentlabels{%
    \answer{}%
  }%
}
&
\topcell{%
  \eliteP
}
\\

\addlinespace[3pt]

\topcell{%
  \texttt{n07}$\to$\texttt{n17}\\[2.5pt]
  \targettag{Adversarial}%
}
&
\topcell{%
  Existence questions were planned primarily around affirmative evidence, causing explicit negations to receive low retrieval priority and making user-level contradictions easy to miss.%
}
&
\topcell{%
  Make existence-query planning contrastive by retrieving both affirmative and negative evidence, and require the answer LLM to surface both sides when they conflict.

  \componentlabels{%
    \plan{}\componentsep%
    \answer{}%
  }%
}
&
\topcell{%
  \eliteA\elitegap
  \eliteT
}
\\

% =========================================================
% Stage III
% =========================================================

\midrule
\multicolumn{4}{@{}l}{\textbf{Stage III: Final integration}} \\

\topcell{%
  \texttt{n20}%
}
&
\topcell{%
  Capability specialists captured complementary strengths, but their mechanisms were not directly composable: aggressive topic grouping, preference enforcement, and aggregation rules could degrade relevance, temporal ordering, or contradiction handling when combined.%
}
&
\multicolumn{2}{
  L{\dimexpr0.525\textwidth-4\tabcolsep\relax}@{}
}{
  \topcell{%
    Merge the complementary mechanisms conservatively by adding topic and standing-preference metadata, lightweight topic-aware scoring, and concise synthesis/preference rules while preserving flat relevance retrieval and adversarial safeguards.

    \componentlabels{%
      \schema{}\componentsep%
      \extract{}\componentsep%
      \indexc{}\componentsep%
      \retrievec{}\componentsep%
      \answer{}\componentsep%
      \helpers{}%
    }%
  }
}
\\

\end{longtable}
\endgroup

\section{Cases and Suggested Revisions on Different Capabilities}
\label{sec:capability_revision_cases}

We examine three cases from memory evolution trajectories on BEAM, each contrasting two revisions with different capability effects and a third revision that addresses both needs. These cases show that different capabilities may suggest different revisions, yet they are not inherently incompatible: by comparing the strengths and limitations of revisions across capabilities, integration can identify and combine complementary improvements. \method{} leverages evolved programs and their evaluation results across capabilities to guide this integration and improve overall performance.

\subsection{Case 1: Store each interaction as a whole versus each fact separately.}
\textit{Temporal tracking} questions compare events or values at different points in a user's history, requiring the memory to preserve relationships among related details. The diagnosis found that a single request or calculation was split across multiple memory units, causing its steps, values, and outcome to be treated as separate interactions. The model then revised the extraction instruction to keep each complete interaction in one unit. This improved temporal tracking by preserving relationships across events. However, grouping related information also combined target statements with long, less informative context, causing the answer LLM to overlook contradictions with opposing statements elsewhere in the history. Temporal tracking increased by 3.41 pp, while adversarial handling decreased by 14.06 pp.

\textit{Adversarial questions} require identifying conflicting claims, such as a user stating both that they have practiced a topic and that they have never practiced it. The diagnosis found that distinct factual statements were sometimes omitted during extraction, motivating a revision to the extraction instruction that preserves each distinct fact independently and recovers missing denials in adversarial questions. However, separating facts also dispersed related values across memory units, making it harder to associate them with the events that produced them. Adversarial handling increased by 12.89 pp, while temporal tracking decreased by 6.90 pp.

These cases show that temporal tracking benefits from preserving relationships among events and values, while contradiction detection requires opposing claims to remain independently accessible. One can refine the extraction instruction to ask the model to \textit{preserve a complete plan, process, or timeline within one memory unit while storing contradictory claims separately.} For example, a problem count and its associated score remain together, while a denial of prior practice remains separate from evidence of completed practice. This revision improved both capabilities: temporal tracking increased by 0.87 pp, and adversarial handling increased by 2.73 pp.

\subsection{Case 2: Reconstruct chronological order versus require explicit causal links.}
\textit{Temporal tracking} questions require reconstructing the order of events across conversations. The diagnosis found that answers grouped concrete events into broad activity categories, obscuring their sequence. The model revised the answer instruction to organize specific events in chronological session order, making the sequence clearer. However, this sometimes led the model to infer a causal relationship from temporal order, treating an event that occurred earlier as having caused a later event. This caused adversarial performance to decrease by 9.38 pp, while temporal tracking increased by 5.61 pp.

\textit{Adversarial} questions about causal influence require checking whether the history explicitly connects a stated cause to its effect. The diagnosis found that chronological ordering of related events was sometimes treated as evidence of causal influence. The model revised the answer instruction to require an explicit causal link and abstain when such evidence was absent. However, this overly strict criterion caused some temporal questions requiring an explicitly stated sequence to be incorrectly rejected. Adversarial handling increased by 8.20 pp, while temporal tracking decreased by 4.25 pp.

These cases highlight two distinct requirements for relating events: temporal tracking can reconstruct order from dated records, whereas causal influence requires explicit evidence connecting cause and effect. A good answer instruction revision should ask the model to \textit{organize answers to temporal and progress questions around the user's specific events in chronological session order, while checking for an explicit connection between cause and effect in causal and influence questions.} This revision improved both capabilities: temporal tracking increased by 2.71 pp, and adversarial handling increased by 8.59 pp.

\subsection{Case 3: Increase retrieval diversity versus tighten relevance filtering.}
\textit{Multi-session synthesis} questions require covering different relevant aspects of a user's history. The diagnosis found that retrieval repeatedly selected similar records from one dominant topic, leaving other aspects underrepresented. The model therefore revised retrieval to use maximal marginal relevance (MMR), balancing relevance to the question with diversity among selected records. This broadened the evidence available for synthesis, but could also introduce records from other events, making it harder to associate dates and updates with the target event in temporal questions. Thus, multi-session synthesis increased by 7.08 pp, while temporal tracking decreased by 5.95 pp.

\textit{Temporal tracking} questions instead require closely related records about the target event to compare dates and track updates. The diagnosis found that records about nearby events interfered with identifying the correct timeline. The model raised the candidate relevance threshold for MMR (e.g., 0.6 $\rightarrow$ 0.7), focusing retrieval on records more closely related to the question. However, stricter filtering also removed records covering other relevant aspects, narrowing the evidence available for broad synthesis. Temporal tracking therefore increased by 5.87 pp, while multi-session synthesis decreased by 5.62 pp.

These cases highlight two retrieval requirements: broad coverage for multi-session synthesis and focused retrieval of related records for temporal tracking. A good retrieval revision should be able to \textit{adjust the MMR balance according to the predicted question type: synthesis questions favor greater diversity (e.g., 0.6 MMR weight), while temporal and progression questions favor relevance to retain closely related records (e.g., 0.7 MMR weight).} This revision improved both capabilities: multi-session synthesis increased by 3.92 pp, and temporal tracking increased by 3.70 pp.

\section{Prompt Catalog}
\label{sec:prompts}

We present the reflection, refinement coder, planning, and integration coder prompts used in \method{}. The reflection and refinement coder templates are used in Stage I (cold start) and Stage II (capability refinement), while the planning and integration coder templates are used in Stage III (integration).

\begingroup
\lstset{basicstyle=\ttfamily,breaklines=true,breakatwhitespace=false,
  columns=fullflexible,keepspaces=true,showstringspaces=false,tabsize=4}
\newtcolorbox{promptstage}[1]{enhanced,colback=white,
  colframe=accent!45!black,colbacktitle=accent!10,coltitle=ink,
  fonttitle=\bfseries,boxrule=0.4pt,arc=1.5pt,before upper=\raggedright,
  left=5pt,right=5pt,top=4pt,bottom=4pt,title={#1}}

\begin{promptcard}{Reflection prompt template used in Stages I and II}
\promptsection{System.}
You are an expert in long-term memory systems for LLM agents. Think carefully, ground every conclusion in the supplied implementation and measurements, and return exactly the requested format.

\promptsection{Task.}
You are diagnosing a memory system serving many long dialogues between a user and an assistant. The system uses an external LLM to generate multiple MemoryUnits from each dialogue segment. It passes all units from each source session and the raw session text with timestamp to Memory.index in chronological order, building memory incrementally across sessions. For each question, the external LLM generates a typed Query, which is passed to Memory.retrieve once. The retrieved text and AnswerInstruction are then provided to the answer LLM. The goal is to maximize answer accuracy within a 16000-character retrieval budget. Consider the recall-precision trade-off throughout the entire memory pipeline: each memory unit should retain sufficient factual information while avoiding unnecessary noise. Too little retrieval may miss key evidence, while too much may interfere with answering.

The memory system is evaluated across five capabilities:
\begin{itemize}[leftmargin=*]
\item \texttt{factual\_retrieval}: preserve and locate explicit facts, names, exact values, lists, and local details;
\item \texttt{temporal}: distinguish states of the same fact at different times, order events, determine recency, and track updates over time;
\item \texttt{preference\_extraction}: recover and apply long-term user preferences, instructions, constraints, and needs;
\item \texttt{multi\_session\_synthesis}: connect and summarize evidence distributed across multiple sessions;
\item \texttt{adversarial}: recognize when the dialogue contains no relevant information and abstain, or detect and cautiously handle contradictory or misleading evidence.
\end{itemize}

Different capabilities may require different memory designs. You can use the Query Planning stage to identify the capabilities required by the question and, when appropriate, adapt the query and retrieval strategy accordingly to preserve accuracy across capabilities.

\noindent\begin{minipage}[t]{0.48\linewidth}\vspace{0pt}
\begin{promptstage}{Stage I: cold start}
Analyze the provided measurements and cases carefully. Identify the capability with the largest accuracy gap, diagnose its most important cause in the current implementation, and propose exactly one diagnosis and implementation change.
\end{promptstage}
\end{minipage}\hfill
\begin{minipage}[t]{0.48\linewidth}\vspace{0pt}
\begin{promptstage}{Stage II: capability refinement}
For the target capability, diagnose the main cause of errors in the current implementation and propose exactly one diagnosis and implementation change.

Target capability: [Target capability]
\end{promptstage}
\end{minipage}\par\medskip

Consider both dimensions before choosing that single change:

(A) \textbf{Prompt Optimization}: How should MemoryExtractionInstruction, QueryPlanningInstruction, or AnswerInstruction change to guide the task LLM better?

(B) \textbf{Memory Design}: How should the MemoryUnit/Query schemas or Memory.index/retrieve logic change to store and retrieve more useful information?

\promptsection{Memory abstractions.} \emph{Insert the memory-interface prompt.}

\promptsection{Current implementation.} [Complete parent source]

\promptsection{Current measurements.}
Training accuracy by capability and average score:

[Training capability profile and average score]

\begin{promptstage}{Stage I only: evidence-based metrics}
Below are diagnostic metrics computed using semantic similarity at different memory stages. The table below reports the average answer accuracy (Acc) and evidence-similarity diagnostics for each capability. Use these metrics to help locate problems in the current implementation. Semantic similarity is robust to paraphrasing, but may remain high even when exact details are missing or changed. These metrics are imperfect intermediate indicators, so analyze them together with representative cases rather than optimizing them directly.
\begin{itemize}[leftmargin=*]
\item \texttt{evidence\_extracted}: how closely golden evidence is represented by generated MemoryUnits from its source segment
\item \texttt{evidence\_oracle\_retrieved}: how reachable golden evidence is through Query generation and Memory.retrieve when the evidence itself is used as the question
\item \texttt{evidence\_retrieved}: how much golden evidence Memory.retrieve returns for the real question
\end{itemize}
[Table with columns: Capability, Acc, evidence\_extracted, evidence\_oracle\_retrieved, evidence\_retrieved; one row per capability]
\end{promptstage}
\medskip

Below are the length-diagnostic measurements across all questions.
\begin{itemize}[leftmargin=*]
\item \texttt{chars\_per\_unit}: the average generated MemoryUnit JSON length; larger and smaller values each have recall and precision trade-offs
\item \texttt{retrieved\_chars}: the average real-question retrieval length; larger and smaller values each have recall and precision trade-offs
\end{itemize}
[Length measurements with name, definition, and current value]

\promptsection{Representative cases.}
Below are representative cases for reference. ``\texttt{diagnostics}'' contains the measurement results for each case.
``\texttt{primary}'' is the primary capability; ``\texttt{secondary}'' lists the secondary capabilities.
``\texttt{original\_text}'' shows the original raw text of each golden evidence.
``\texttt{extracted\_as}'' shows how golden evidence text is extracted and stored in MemoryUnits by the current implementation.
``\texttt{retrieved}'' shows text returned by Memory.retrieve for the question.

If ``\texttt{original\_text}'', ``\texttt{extracted\_as}'', or ``\texttt{retrieved}'' is too long, it may be truncated to retain information relevant to answering the question.

[Representative cases serialized as JSON: one per capability in Stage I, or up to five target-capability cases in Stage II]

\promptsection{Previously explored revisions (when available).}

\noindent\begin{minipage}[t]{0.48\linewidth}\vspace{0pt}
\begin{promptstage}{Stage I: cold start}
Here are previously explored revisions from the same current implementation. You can refer to them and avoid repeating unhelpful revision directions.

[Base and child revisions with summaries, modified components, capability scores, and average score]
\end{promptstage}
\end{minipage}\hfill
\begin{minipage}[t]{0.48\linewidth}\vspace{0pt}
\begin{promptstage}{Stage II: capability refinement}
Here are previously committed versions for the target capability. Use them to understand which changes have already been explored and avoid reverting useful improvements.

[JSON list with node, summary, components, and training\_profile]
\end{promptstage}
\end{minipage}\par\medskip

\promptsection{Output.}
Analyze the provided measurements and cases to identify the main reasons why the current implementation fails to produce correct results. Ground the diagnosis in the implementation, measurements, previously explored revisions, repeated nearby unsuccessful directions when shown, and representative cases. Return one diagnosis with one concrete implementation change.

Answer with exactly one JSON object:
\begin{lstlisting}
{
  "diagnosis": "<one code-and-case-grounded problem and concrete implementation change>"
}
\end{lstlisting}
\end{promptcard}

\begin{promptcard}{Refinement Coder prompt template used in Stages I and II}
\promptsection{System.}
You are an expert in long-term memory systems for LLM agents. Think carefully, ground every conclusion in the supplied implementation and measurements, and return exactly the requested format.

\promptsection{Task.}
You are improving a memory system serving many long dialogues between a user and an assistant. The system uses an external LLM to generate multiple MemoryUnits from each dialogue segment. It passes all units from each source session and the raw session text with timestamp to Memory.index in chronological order, building memory incrementally across sessions. For each question, the external LLM generates a typed Query, which is passed to Memory.retrieve once. The retrieved text and AnswerInstruction are then provided to the answer LLM. The goal is to maximize answer accuracy within a 16000-character retrieval budget. Consider the recall-precision trade-off throughout the entire memory pipeline: each memory unit should retain sufficient factual information while avoiding unnecessary noise. Too little retrieval may miss key evidence, while too much may interfere with answering.

The test questions cover temporal reasoning, evidence connected across sessions, facts updated over time, contradictions that must be surfaced, abstention when dialogue contains no answer, and long-term preferences or instructions that should guide an answer.

Implement the requested change based on the diagnosis feedback for the current implementation; do not broaden it into an unrelated redesign. Implement the requested algorithm efficiently without sacrificing its intended functionality or accuracy.

\promptsection{Memory abstractions.} \emph{Insert the memory-interface prompt.}

\promptsection{Current implementation.} [Complete fixed parent source]

\promptsection{Diagnosis.} [Diagnosis]

\promptsection{Representative cases.}
Use the following cases to understand the diagnosis and implement the requested change.
``\texttt{primary}'' is the primary capability; ``\texttt{secondary}'' lists the secondary capabilities.
``\texttt{original\_text}'' shows the original raw text of each golden evidence.
``\texttt{extracted\_as}'' shows how golden evidence text is extracted and stored in MemoryUnits by the current implementation.
``\texttt{retrieved}'' shows text returned by Memory.retrieve for the question.

If ``\texttt{original\_text}'', ``\texttt{extracted\_as}'', or ``\texttt{retrieved}'' is too long, it may be truncated to retain information relevant to answering the question.

[The same compact cases used for diagnosis, serialized as JSON]

\promptsection{Revision history (when available).}

\noindent\begin{minipage}[t]{0.48\linewidth}\vspace{0pt}
\begin{promptstage}{Stage I: cold start}
The following are recent, similar revision directions that did not improve accuracy. Do not repeat these directions. Use the diagnosis to make a change in a different component or another part of the memory pipeline.

[Revision summaries and modified components]
\end{promptstage}
\end{minipage}\hfill
\begin{minipage}[t]{0.48\linewidth}\vspace{0pt}
\begin{promptstage}{Stage II: capability refinement}
Here are previously committed versions for the target capability. Use them to understand which changes have already been explored and avoid reverting useful improvements.

[JSON list with node, summary, components, and training\_profile]
\end{promptstage}
\end{minipage}\par\medskip

\promptsection{How to write the patch.}
The patcher locates hunks from quoted context rather than trusting line numbers. Copy context exactly, begin every changed region with an \texttt{@@} hunk header, and use separate hunks for non-contiguous edits. The patch must compile, make a real change, and preserve the required schemas, instructions, and three Memory signatures. Return the diff against Current implementation, not against a previous attempt.

Example:
\begin{lstlisting}
@@
-MemoryExtractionInstruction = "Extract the key information from the text into several memory units."
+MemoryExtractionInstruction = "Extract distinct facts, updates, and preferences from the text into several memory units."
\end{lstlisting}

\promptsection{Previous-attempt feedback (when available).}

\emph{If the previous attempt failed during patching, checks, or execution:}

Previous failed attempt:

[Previous summary, code, and error]

Repair this error. Return a complete diff against the original Current implementation.

\emph{If the previous attempt was measured but not accepted:}

Previous measured attempt:

[Previous reply JSON with summary and code]

[Measurement JSON with parent, candidate, and reasons]

Analyze this failure and revise the original parent following the diagnosis.

\promptsection{Output.}
Answer with exactly one JSON object:
\begin{lstlisting}
{
  "summary": "<one concise change description>",
  "code": "<unified diff against Current implementation>"
}
\end{lstlisting}
\end{promptcard}

\begin{promptcard}{Planning prompt template used in Stage III}
\promptsection{System.}
You are an expert in long-term memory systems for LLM agents. Think carefully, ground every conclusion in the supplied implementation and measurements, and return exactly the requested format.

\promptsection{Task.}
Plan one integrated Memory Program using Current implementation as the sole base. Compare the supplied capability specialists using their implementations, training gains over the base, and paired cases. Select compatible improvements, retain useful base behavior, and resolve conflicts and shared dependencies.
Return an implementation-ready plan: specify what to change, where, and how, including required schema, instruction, indexing, retrieval, and helper updates.

\promptsection{Memory abstractions.} \emph{Insert the memory-interface prompt.}

\promptsection{Current implementation.} [Complete overall-best base source]

\promptsection{Capability specialist sources.}
[JSON object identifying base-equivalent versions and distinct specialist sources, with their capability roles and training gains over the base]

\promptsection{Current measurements.}
Training accuracy by capability and average score:

[Base capability-profile JSON]

Length diagnostics:
\begin{itemize}[leftmargin=*]
\item \texttt{chars\_per\_unit}: the average generated MemoryUnit JSON length; larger and smaller values each have recall and precision trade-offs
\item \texttt{retrieved\_chars}: the average real-question retrieval length; larger and smaller values each have recall and precision trade-offs
\end{itemize}
[JSON list with name, definition, and value for each length diagnostic]

\promptsection{Representative cases.}
Compare base with capability specialist outputs; \texttt{*\_parent} denotes a specialist's earlier parent for retention comparison.
``\texttt{original\_text}'' shows the original raw text of each golden evidence.
``\texttt{extracted\_as}'' shows how golden evidence text is extracted and stored in MemoryUnits by the current implementation.
``\texttt{retrieved}'' shows text returned by Memory.retrieve for the question.

If ``\texttt{original\_text}'', ``\texttt{extracted\_as}'', or ``\texttt{retrieved}'' is too long, it may be truncated to retain information relevant to answering the question.

[Paired-case JSON: question, golden\_answer, original\_text, and the available base, capability\_case, and capability\_parent outputs; each output contains prediction\_answer, answer\_score, query\_plan, retrieved, and extracted\_as]

\promptsection{Output.}
Return exactly one JSON object:
\begin{lstlisting}
{"Step-by-step plan": "<concrete integration plan specifying changes and dependencies>"}
\end{lstlisting}
\end{promptcard}

\begin{promptcard}{Integration Coder prompt template used in Stage III}
\promptsection{System.}
You are an expert in long-term memory systems for LLM agents. Think carefully, ground every conclusion in the supplied implementation and measurements, and return exactly the requested format.

\promptsection{Task.}
Implement the integration plan against Current implementation, using the supplied capability specialist sources. Integrate required schema, instruction, indexing, retrieval, and helper dependencies together; resolve conflicts and preserve compatible base behavior. Use measured case changes to refine the integration, not benchmark-specific rules.

\promptsection{Memory abstractions.} \emph{Insert the memory-interface prompt.}

\promptsection{Current implementation.} [Complete fixed overall-best base source]

\promptsection{Capability specialist sources.}
[JSON list of distinct specialist sources, their capability\_case version names, and training\_gain\_over\_base]

\promptsection{Merge plan.} [JSON object containing the integration plan]

\promptsection{Representative cases.}
``\texttt{original\_text}'' shows the original raw text of each golden evidence.
``\texttt{extracted\_as}'' shows how golden evidence text is extracted and stored in MemoryUnits by the current implementation.
``\texttt{retrieved}'' shows text returned by Memory.retrieve for the question.

If ``\texttt{original\_text}'', ``\texttt{extracted\_as}'', or ``\texttt{retrieved}'' is too long, it may be truncated to retain information relevant to answering the question.

[Paired-case JSON with the shared question, golden answer, original evidence, and base and capability specialist outputs]

\promptsection{How to write the patch.}
The patcher locates hunks from quoted context rather than trusting line numbers. Copy context exactly, begin every changed region with an \texttt{@@} hunk header, and use separate hunks for non-contiguous edits. The patch must compile, make a real change, and preserve the required schemas, instructions, and three Memory signatures. Return the diff against Current implementation, not against a previous attempt.

Example:
\begin{lstlisting}
@@
-MemoryExtractionInstruction = "Extract the key information from the text into several memory units."
+MemoryExtractionInstruction = "Extract distinct facts, updates, and preferences from the text into several memory units."
\end{lstlisting}

\promptsection{Previous attempts (when available).}
Compare each attempt's case outputs with the corresponding base and capability specialist outputs. Use the implementations to identify integration gains, conflicts, and regressions. Missing views are unavailable, not unchanged. Return a complete diff against Current implementation.

[JSON object keyed by base\_attempt1, base\_attempt2, etc.; each entry contains the available source or patch, training result, error, and per-case candidate outputs]

\promptsection{Output.}
Return exactly one JSON object:
\begin{lstlisting}
{
  "summary": "<one concise change description>",
  "code": "<unified diff against Current implementation>"
}
\end{lstlisting}
\end{promptcard}
\endgroup

\section{Evolved Memory Program}
\label{sec:final_programs}

In this section, we present the final memory programs evolved by \method{} on BEAM and LongMemEval.

\subsection{BEAM Final Memory Program}
The program evolved by \method{} on BEAM augments dated atomic memories with topic identifiers and a flag for broadly applicable user preferences or standing instructions.
\begin{itemize}[leftmargin=*]
  \item Extraction assigns related facts to a shared topic and distinguishes cross-topic preferences from project-specific constraints.
  \item Indexing stores the embeddings and metadata, recovers missing session dates from the input header, and collects distinct standing preferences separately.
  \item Planning splits multi-aspect questions into subqueries and adds a query for negative evidence when checking whether an action occurred.
  \item Each subquery ranks memory units using cosine similarity with weight 0.7 and lexical query-token coverage with weight 0.3, plus a topic-based bonus of 0.05. Within the 16000-character context budget, retrieval reserves space for a standing-preference block capped at 400 characters and divides the remaining space equally among subqueries. It removes exact-text duplicates across subqueries and appends the preference block after the date-prefixed results.
  \item The answering instruction distinguishes statement time from event time, organizes chronological summaries, separates distinct-event sums from running totals, and applies retrieved preferences to the response. It calls for clarification when affirmative and negative evidence conflict, and abstention when evidence is insufficient.
\end{itemize}

\tcbinputlisting{papercard,colframe=green!40!black,colbacktitle=green!40!black,
  title={Final memory program evolved by \method{} for BEAM},
  listing only,listing file={programs/beam_final.py},
  listing options={language=Python,basicstyle=\ttfamily\small,breaklines=true,
    breakatwhitespace=false,columns=fullflexible,keepspaces=true,
    showstringspaces=false,tabsize=4}}

\subsection{LongMemEval Final Memory Program}
The program evolved by \method{} on LongMemEval uses a flat dense index of dated atomic memories.
\begin{itemize}[leftmargin=*]
  \item Extraction preserves exact names, values, dates, lists, and user preferences, resolves references to explicit entity names, and records the source session date.
  \item Indexing stores each unit's text, normalized embedding, and date in aligned lists.
  \item Planning generates a single query tailored to the question's entities, factual qualifiers, time references, or preference cues.
  \item Retrieval ranks all memory units by cosine similarity, keeps those scoring at least 0.25, and packs date-prefixed units in relevance order within a 16000-character context budget.
  \item The answering instruction applies remembered constraints to recommendations, uses date prefixes for temporal calculations, enforces factual qualifiers, aggregates distinct occurrences, and requests abstention when a specific fact is unsupported.
\end{itemize}

\tcbinputlisting{papercard,colframe=green!40!black,colbacktitle=green!40!black,
  title={Final memory program evolved by \method{} for LongMemEval},
  listing only,listing file={programs/longmemeval_final.py},
  listing options={language=Python,basicstyle=\ttfamily\small,breaklines=true,
    breakatwhitespace=false,columns=fullflexible,keepspaces=true,
    showstringspaces=false,tabsize=4}}